\documentclass[a4paper]{article}
\usepackage[text={7in,10in},centering]{geometry}
\usepackage{graphicx,color}
\usepackage{amsmath,amsthm,amssymb}
\usepackage{multirow,array}
\usepackage[table]{xcolor}
\usepackage{caption,subcaption}
\usepackage{url}
\usepackage{algorithmic} 
\usepackage[ruled,vlined]{algorithm2e}
\usepackage{hyperref}
\usepackage{cleveref}

\newcommand{\eg}{{\it e.g.}}
\newcommand{\ie}{{\it i.e.}}
\newcommand{\reals}{{\mbox{\bf R}}}
\newcommand{\Cov}{\mathop{\bf cov{}}}
\newcommand{\minimize}{\mathop{\rm minimize{}}}
\newcommand{\argmin}{\mathop{\rm argmin}}
\newcommand{\prox}{\mbox{\bf prox}}
\newcommand{\nullspace}{{\mathcal N}}
\newcommand{\Range}{\mathop{\bf range}}

\graphicspath{{figures/}}

\title{Joint learning of multiple Granger causal networks \\ via non-convex regularizations: Inference of group-level brain connectivity}

\author{Parinthorn Manomaisaowapak and Jitkomut Songsiri\footnote{Corresponding author} \\[1ex]
 Department of Electrical Engineering, Faculty of Engineering \\ [1ex]
 Chulalongkorn University, Bangkok, Thailand 10330 \\[1ex]
 e-mail: parinthorn@gmail.com, and jitkomut.s@chula.ac.th
}   

\begin{document}
\maketitle

\begin{abstract}
This paper considers joint learning of multiple sparse Granger graphical models to discover underlying common and differential Granger causality (GC) structures across multiple time series. This can be applied to drawing group-level brain connectivity inferences from a homogeneous group of subjects or discovering network differences among groups of signals collected under heterogenous conditions. By recognizing that the GC of a single multivariate time series can be characterized by common zeros of vector autoregressive (VAR) lag coefficients, a group sparse prior is included in joint regularized least-squares estimations of multiple VAR models. Group-norm regularizations based on group- and fused-lasso penalties encourage a decomposition of multiple networks into a common GC structure, with other remaining parts defined in individual-specific networks. Prior information about sparseness and sparsity patterns of desired GC networks are incorporated as relative weights, while a non-convex group norm in the penalty is proposed to enhance the accuracy of network estimation in low-sample settings. Extensive numerical results on simulations illustrated our method's improvements over existing sparse estimation approaches on GC network sparsity recovery. Our methods were also applied to available resting-state fMRI time series from the ADHD-200 data sets to learn the differences of causality mechanisms, called effective brain connectivity, between adolescents with ADHD and typically developing children.  Our analysis revealed that parts of the causality differences between the two groups often resided in the orbitofrontal region and areas associated with the limbic system, which agreed with clinical findings and data-driven results in previous studies. 
\end{abstract}


\section{Introduction}
\label{sec:intro}

The influences exerted by one region of the human brain on another is described as the effective brain connectivity, or the causality flow within the human brain. Granger causality (GC) analysis is a model-based method that can reveal effective brain connectivity or causal interconnections among brain regions from neural activity data using vector autoregressive (VAR) models; see a review of connectivity inference in \cite{AYD:18}. Unfortunately, without prior information on the estimated network of interconnections, the resulting Granger graphical model (or GC network) is typically a dense estimate. Since it is difficult to make inferences about the brain's interactive structure from dense models, sparser solutions are desirable, which can be achieved using regularization and Bayesian inference frameworks \cite{AYD:18}. Sparse GC networks are studied in many variations for insignificant causality filtering, including lasso GC~\cite{FS07}, group lasso GC~\cite{LALR09}, or truncated lasso~\cite{SM10}. Recently, \cite{BLHLYU20} proposed to use a non-convex $\ell_{q}$ penalty with $0<q<1$ to improve GC estimation since it was theoretically superior to a lasso-type penalty in a linear regression context \cite{WCLQ18}.

This paper investigates advancing GC network estimations in two ways: simultaneously learning multiple GC networks, and constructing GC networks using a non-convex sparsity-inducing penalty. Simultaneous learning in the context of multiple sparse Gaussian graphical models currently exists as an estimation of the sparse inverse covariance matrix of random vectors, otherwise known as the graphical lasso problem \cite{FHT07}. A zero pattern of the inverse covariance, also known as the precision matrix, indicates a conditional independence between random variables. This framework has been extended to joint Gaussian graphical model estimations with added prior knowledge on possible relationships among all models to force the models decomposed into two parts, a common and a differential part. The common part is defined as the common nonzeros in the inverse covariance matrices of all models, and the differential part refers to all other remaining nonzeros. Prior knowledge of common causality connections can be used to form a group lasso regularization to make all models have identical sparsity patterns \cite{LCC16,MM16}, or a fused lasso to shrink the parameter differences among models \cite{DWW14,SS16}. \cite{HaW:13} decomposed the precision matrices of multiple models into a common part whose sparsity was promoted by a lasso, and individual parts that were jointly regularized through a convex composite $\ell_{1,p}$ norm\footnote{The notation of $\ell_{p,q}$ refers to the composite of $\ell_q$ with $\ell_p$ norm. We note that when $q < 1$, $\Vert \cdot \Vert_q$ is not strictly a norm.} for $p=1,2,\infty$. Other variants of single Gaussian graphical modelling approaches extended for multiple modeling also exploited similar lasso-type techniques; for example, these have involved a row and column inverse covariance estimation of the matrix Gaussian distribution  \cite{HC14}, or the estimation of the inverse covariance and correlation matrices of Gaussian graphical models with explanatory variables \cite{HCH18}.

Joint learning of Gaussian graphs has also been extended to use non-convex penalties, such as $\ell_{0}$ to regularize non-zero elements~\cite{THW16}, the $\ell_{1,1/2}$ norm \cite{GLM11}, a general composite $\ell_{1,q}$ norm with $0 < q < 1$, truncate logarithmic penalty and inverse polynomial penalty~\cite{CZZ15}. Despite several extensions of Gaussian graphical models to joint modeling, Gaussian graphs lack the core functionalities of capturing the temporal dependencies of time series and causality directions, issues which can solved by multiple sparse GC modeling.

Current approaches to jointly estimate multiple GC graphs have relied on VAR models extracting the common edges among all estimated networks then placing them into \emph{a common GC network}, and the differential edges that are unique to individual models then organizing them as \emph{differential GC networks}. A group lasso is often applied to penalize parameters across models in the estimation of a common GC network \cite{JSSunpub, G15}. To learn both common and differential GC networks, several studies employed a combination of fused lasso and other lasso variants in regularized least-squares estimations \cite{SkM:19a,WBC16,Son:15}. The fused lasso encouraged some parameter entries to be identical across models, establishing the common network, while the other lasso-type penalty promoted zeros in each model's parameters in building the differential networks. However, because the VAR-based null GC of a single model contains all the \emph{common zeros of all lag coefficients}~\cite{Lut:05}, when the lasso-type regularization did not penalize all VAR-lag parameters as in \cite{SkM:19a,WBC16}, the induced sparsity pattern did not directly reflect the null GC. Instead, VAR-lag matrices of each single model should be penalized in groups, as was accomplished with a group-norm penalty in \cite{Son:15,SkM:19b,G15}. Finally, \cite{SkM:19b} proposed a two-stage procedure to jointly estimate multiple GC graphs using a group lasso penalty where a common and differential GC networks were extracted in two separate stages, respectively. 

Meanwhile, all of the aforementioned studies have relied on group- and fused-lasso which are convex penalties. Applying non-convex penalties in applications of GC analysis is a relatively new approach, and only two recent works are worth noting. In one study, \cite{BLHLYU20} proposed using the non-convex $\ell_{1/2}$-norm penalty for estimating a single GC network, but without a group penalization of all VAR-lag parameters to relate null GC with the zero group of parameters; this group-norm penalty ($\ell_{2,1/2}$) was an extension from the $\ell_{1/2}$ norm which had been shown to yield superior performances in recovering the true sparse solution over its convex counterpart~\cite{HLMQY17}. In the other study, the $\ell_{2,1/2}$ penalty was applied in~\cite{MS20} to estimate multiple VAR models having an identical GC structure. Despite a performance gain from the non-convex penalty under some settings, the formulation in \cite{MS20} have room for improvement by introducing relative weights in the group penalty as a prior describing the edge strengths in the GC networks. This can be accomplished by adopting the adaptive group lasso in~\cite{WL08} that allowed different amounts of shrinkage on regression coefficients which saw improved variable selection performances over the lasso.

From the literature, we can draw on the strengths of various relevant methods and craft stronger regularization techniques consisting of three main features: i) use the group-norm penalization of individual model parameters to correctly infer null GC from the estimated group-sparse structure, ii) use the non-convex $\ell_{2,1/2}$ penalty to enhance the recovery rate of true parameters, and iii) use relative weights with group penalties to improve the accuracy of detecting edges in an estimated network. All three features, applied to jointly estimate multiple GC networks, form the contributions of this paper. 

We propose three main formulations that each employ all three features listed above: one estimates multiple GC networks to have the same structure, and the other two estimate multiple models inferring both common and individually specific GC networks. An example application of revealing a common GC among multiple models is a group-level inference of brain connectivity, where data sets contain signals of several subjects recorded under a controlled condition (\eg, resting-state), and each model parameter belongs to each patient. Presumably, each patient contributes to a homogeneous brain connectivity structure that can be inferred from the estimated common GC network, while the model parameters are allowed to differ according to each patient's profile. As for examples of discovering differential networks, the brain connectivity structure among subjects would be assumed to contain differences, perhaps arising from the testing of patients under two or more conditions (\eg, control versus abnormal brain patterns). In light of these examples, the effectiveness of our three formulations will be illustrated by using the fMRI ADHD-200 data sets to learn the group-level brain network differences between adolescents with attention deficit hyperactivity disorder (ADHD) and typically developing children (TDC). 

\paragraph{Notation} A vector $x$ is partitioned into $m$ blocks as $x=(x_1,x_2,\ldots,x_m)$ and a group-norm $\ell_{p,q}$ of $x$ is defined as $ \Vert x \Vert_{p,q} = (\sum_i \Vert x_i \Vert_p^q)^{1/q}$. We often use the $\ell_{p,q}$ norm to the power of $q$. 

\section{Methods}
\label{section:Methods}
Given $K$ set of vector time series $\{y^{(k)}(t) \}_{t=1}^T$ for $k=1,2,\ldots,K$, denote as the $y^{(k)}_j$ the $j^\mathrm{th}$ component of an $n$-dimensional vector time series obtained from the $k^\mathrm{th}$ data set where $j=1,\ldots,n$ and $k=1,\ldots,K$. We fit each $y^{(k)}(t)$ to a $p$-order VAR model described by 
\begin{equation}
y^{(k)}(t) = \sum_{r=1}^{p}A^{(k)}_{r}y^{(k)}(t-r) + \epsilon^{(k)}(t), \;\; k=1,\ldots,K,
\label{eq:var}
\end{equation}
where $y^{(k)} \in \reals^{n}$, $A_{r}^{(k)} \in \reals^{n\times n}$ for $r =1,2,\ldots,p$,  and $\epsilon^{(k)} \in \reals^{n}$ is assumed to be a white noise. The $(i,j)$ entry of $A_{r}^{(k)}$, denoted as $(A_{r}^{(k)})_{ij}$, is a coupling gain from $y_j^{(k)}$ at lag $r$ to $y^{(k)}_i$. Directly from~\cite{Lut:05}, when considering the $k^\mathrm{th}$ model, there is no Granger causality from $y_j^{(k)}$ to $y^{(k)}_i$ if and only if 
\begin{equation}
B^{(k)}_{ij}=\begin{bmatrix}
(A^{(k)}_{1})_{ij} & \cdots & (A^{(k)}_{p})_{ij}
\end{bmatrix} \in \reals^p
\label{eq:bij}
\end{equation}
is a zero vector (\ie, $(A^{(k)}_r)_{ij}$ is zero \emph{for all} $r$). Hence, a joint estimation of multiple sparse Granger graphical models is typically formulated in a least-square sense with a sparse-inducing penalty as
\begin{equation}
\underset{A^{(1)},\ldots,A^{(K)}}{\minimize} ~ \frac{1}{2N}\sum_{k=1}^{K} \left \Vert Y^{(k)}-A^{(k)}H^{(k)} \right \Vert _{F}^{2}+g(A^{(1)},\ldots, A^{(K)}),
\label{eq:jointGC}
\end{equation}
where $\Vert \cdot \Vert_F^2$ is the quadratic loss (Frobenious norm). The problem parameters are $N=T-p$, and $K$ pairs of measurements and the regressor matrix $(Y^{(k)},H^{(k)})$ for $k=1,\ldots,K$. The estimated variables are the VAR coefficients of all $K$ models, $A^{(k)} = \begin{bmatrix} A^{(k)}_{1} & \cdots & A^{(k)}_{p} \end{bmatrix}$, for $k=1,\ldots,K$. The regularization $g$ is constructed for promoting a sparsity in $B_{ij}^{(k)}$ (or equivalently in $A^{(k)}_{ij}$ as described in \eqref{eq:bij}). When there is no relationship priors among the $K$ models, the sparsity of $B^{(k)}_{ij}$ can be regularized independently using a separable penalty function $g$, together with a penalty parameter $\lambda$ as
\begin{equation}
g(A)= \lambda \sum_{k=1}^{K} \sum_{i \neq j}   w_{ij}^{(k)} \Vert B^{(k)}_{ij} \Vert_{2}^{q},
\label{eq:groupnorm}
\end{equation}
where $0< q \leq 1$ and its choice will be specified later. We emphasize on the use of $ \Vert B_{ij}^{(k)} \Vert_2^q$ instead of the conventional lasso of VAR coefficients directly as $\sum_{r=1}^p | (A^{(k)}_r)_{ij}|$ in \cite{SkM:19a,WBC16} since the lasso does not promote a \emph{common zero} among \emph{all VAR-lag} parameters; see the original GC definition in \eqref{eq:bij}. For a fixed $k$, a pre-defined $w_{ij}^{(k)} > 0$ indicates the degree to which each $B_{ij}^{(k)}$ is penalized, or equivalently, it gives a likelihood prior of GC from $y_j^{(k)}$ to $y_i^{(k)}$. The joint sparse estimation \eqref{eq:jointGC} with $g$ in \eqref{eq:groupnorm} can then be optimized separately for each $k$. The topology of the $k^\mathrm{th}$ model's GC network is specified by the sparsity pattern of an $n \times n$ matrix formed by the estimated $\Vert B^{(k)}_{ij} \Vert_2$ for $1 \leq i,j \leq n$.

We predicate three kinds of GC relationships among the models. The first proposal restricts all models to have the same sparse GC-network pattern, which is made possible using the notion of group norm penalties such as the $\ell_{2,1}$ (group lasso) penalty. The other two formulations account for both potential similarities and differences among the models; this time, the GC networks are decomposed into a common and differential parts. The third formulation adds a restriction that all $K$ models have \emph{identical} VAR coefficients corresponding to their common part; this can be achieved using a fused lasso penalty.

The regularization \eqref{eq:groupnorm} is a group-norm $\ell_{2,q}$ penalty where the two common choices of $q=1$ and $q=1/2$ make problem \eqref{eq:jointGC} into convex and non-convex, respectively. A recovery bound of group-sparse solutions to an $\ell_{p,q}$-regularized regression can be guaranteed upon the $(p,q)$-group restricted eigenvalue condition (GREC) \cite{HLMQY17}\footnote{For $\Vert Ax - b \Vert_2^2 + \lambda \Vert x \Vert_{p,q}$, the condition requires the positive definiteness of $A^TA$ on the associated subblocks. The bound of the estimation error is a big $\mathcal{O}$ of $\lambda$ and the group sparsity level of the true parameter; see Theorem 9 in \cite{HLMQY17}.} with an important property that (2,1)-GREC implies (2,1/2)-GREC. This is a favorable result because the use of an $\ell_{2,1/2}$ penalty requires a condition that is easier to satisfy than $\ell_{2,1}$ to obtain the recovery bound. The experimental results of \cite{HLMQY17} also suggested that the range of true sparsity levels that yielded 100\% successful recovery rate in $\ell_{2,1/2}$ was wider than that of $\ell_{2,1}$. 

With these assumptions on GC network similarities, together with the benefit of the non-convex $\ell_{2,1/2}$ penalty and the choice of penalty weight, we are able to craft penalty functions of our three estimation models as follows.

\subsection{Common GC network estimation}
We propose \textbf{CommonGrangerNet (CGN)} as the formulation for estimating a common GC network of all $K$ models. A common sparsity can be obtained by pooling $B_{ij}^{(k)}$ from all $K$ models into $C_{ij} = \begin{bmatrix} B_{ij}^{(1)} & B_{ij}^{(2)} & \cdots & B_{ij}^{(K)}  \end{bmatrix} \in \reals^{pK}$ and regularizing $C_{ij}$ using the group norm penalty
\begin{equation}
g(A)= \lambda \sum_{i \neq j} v_{ij} \Vert C_{ij} \Vert_{2}^{q}, \quad 0 < q \leq 1.
\label{eq:common_penalty}
\end{equation}
Since the summation of non-negative quantities over $(i,j)$ behaves like an $\ell_1$ penalty, when the penalty parameter $\lambda$ is sufficiently large, some $C_{ij}$'s (from $1 \leq i,j, \leq n$) are zero, revealing a common GC structure among $K$ models. The relative penalty weight among $(i,j)$ is chosen as $v_{ij}=1/\Vert \tilde{C}_{ij} \Vert_{2} ^{q}$ where $\tilde{C}_{ij}$ is the least-squares (LS) estimate of $C_{ij}$; this choice was selected because, if the group norm of the $(i,j)$ entry of all VAR estimates is large, it is likely that $y_j^{(k)}$ has a Granger-cause to $y_i^{(k)}$ for all $k$, and therefore this $(i,j)$ entry should be less penalized. \cite{JSSunpub,G15} previously considered the penalty of $q=1$ in \eqref{eq:common_penalty}, but with an equal amount of shrinkage to all $(i,j)$ by using $v_{ij}=1$. The convex case with $q=1$ also corresponds to the adaptive group lasso in~\cite{WL08}. 

\subsection{Common and differential GC network estimation}
\label{sec:dgn}
We can capture the essence of when all the models are assumed to share a partially similar GC structure but also individually contain unique networks by combining the two regularizations in \eqref{eq:groupnorm} and \eqref{eq:common_penalty} to define
\begin{equation}
g(A)= \lambda_{1} \sum_{k=1}^{K} \sum_{i \neq j}  w_{ij}^{(k)} \Vert B^{(k)}_{ij} \Vert_{2}^{q} + \lambda_2 \sum_{i \neq j} v_{ij} \Vert C_{ij} \Vert_{2}^{q} ,\quad 0 < q \leq 1.
\label{eq:common_diff_penalty}
\end{equation}
In \eqref{eq:common_diff_penalty}, the second term of $g$ promotes a \emph{shared} null GC for \emph{all} GC networks at some $(i,j)$ entries, while the first term regularizes the other $(i,j)$ entries of \emph{individual} models. The nested penalization of both parts encourages a collection of multiple GC networks to simultaneously consist of the common links that are present in all the GC networks, and additional links that vary among individual networks. Similar to \eqref{eq:common_penalty}, the relative weights are chosen as the inverse of the LS estimate: $w_{ij}^{(k)} = 1/\Vert \tilde{B}^{(k)}_{ij} \Vert_2$ and $v_{ij}=1/\Vert \tilde{C}_{ij} \Vert_{2} ^{q}$. The estimation \eqref{eq:jointGC} with penalty \eqref{eq:common_diff_penalty} is referred to as \textbf{DifferentialGrangerNet (DGN)}.

\subsection{Fused and differential GC estimation}
\label{sec:fgn}
Building upon DGN's network structure that contains both common and differential parts, one can force the underlying VAR parameters associated with the common part to be identical among all $K$ models. By replacing the second term of \eqref{eq:common_diff_penalty} with $\Vert B^{(k)}_{ij}-B^{(l)}_{ij} \Vert_{2}^{q} $  where $1 \leq k,l \leq K$ are the model indices, the penalty of \textbf{FusedGrangerNet (FGN)} is
\begin{equation}
g(A)= \lambda_{1}  \sum_{k=1}^{K} \sum_{i \neq j}  w_{ij}^{(k)}  \Vert B^{(k)}_{ij}  \Vert_{2}^{q}
 + \lambda_2 \sum_{k<l}\sum_{i \neq j} u_{ijkl} \Vert B^{(k)}_{ij}-B^{(l)}_{ij} \Vert_{2}^{q} .
\label{eq:sim_diff_penalty}
\end{equation}
The second term of $g$ is a fused lasso that regularizes all possible differences (through the sum $\sum_{k < l}$) between $B^{(k)}_{ij}$ and $B^{(l)}_{ij}$. When the difference is shrunk to zero for some $(i,j)$, the two models will share a common part. The positive weight $u_{ijkl}$ gives a relative degree to the fused lasso's penalization of $(i,j)$ entries and all pairs of model $k$ and $l$. We select $u_{ijkl} = 1/ \Vert \tilde{B}^{(k)}_{ij}-\tilde{B}^{(l)}_{ij} \Vert_{2}^{q}$ where $\tilde{B}_{ij}^{(k)}$ is the LS estimate.

The fused term considered in~\cite{SkM:19a} was constructed differently via \eqref{eq:sim_diff_penalty}, \ie, it was in the form of $\sum_{k < l} \Vert x^{(k)}- x^{(l)} \Vert_2^q$ where $x^{(k)}$ was pooled from \emph{all $(i,j)$ entries} of VAR-lag coefficients of a single model, while our vector $B^{(k)}_{ij}$ was pooled from an $(i,j)$ entry of lag coefficients. Both of these approaches can force two models to share identical parameters due to the fused-lasso feature. However, when $p>1$, we desire estimation results where a single model's parameters describe a sparsity that occurs as a \emph{group} of all VAR-lag coefficients of some $(i,j)$; such zero pattern characteristically infer a causality from variable $j$ to $i$, which can be achieved by \eqref{eq:sim_diff_penalty}, but not~\cite{SkM:19a}. Despite \eqref{eq:sim_diff_penalty}'s characteristic of penalizing any two models in their entirety, the convex fused terms in \cite{Son:15} penalized only two consecutive models as $\Vert B^{(k+1)}_{ij}-B^{(k)}_{ij} \Vert_2$. Also, \cite{Son:15} did not apply prior information about the common part, instead simply using $u_{ijkl} = 1$, while our choice of $u_{ijkl}$ can be made based on which $(i,j)$ entries or pair of two models have more likelihood of having identical parameters.

\subsection{Causality learning scheme}
\label{section:scheme}

\begin{figure}
\centering
\begin{subfigure}[c]{0.4\linewidth}
\centering
\includegraphics[width=0.8\linewidth]{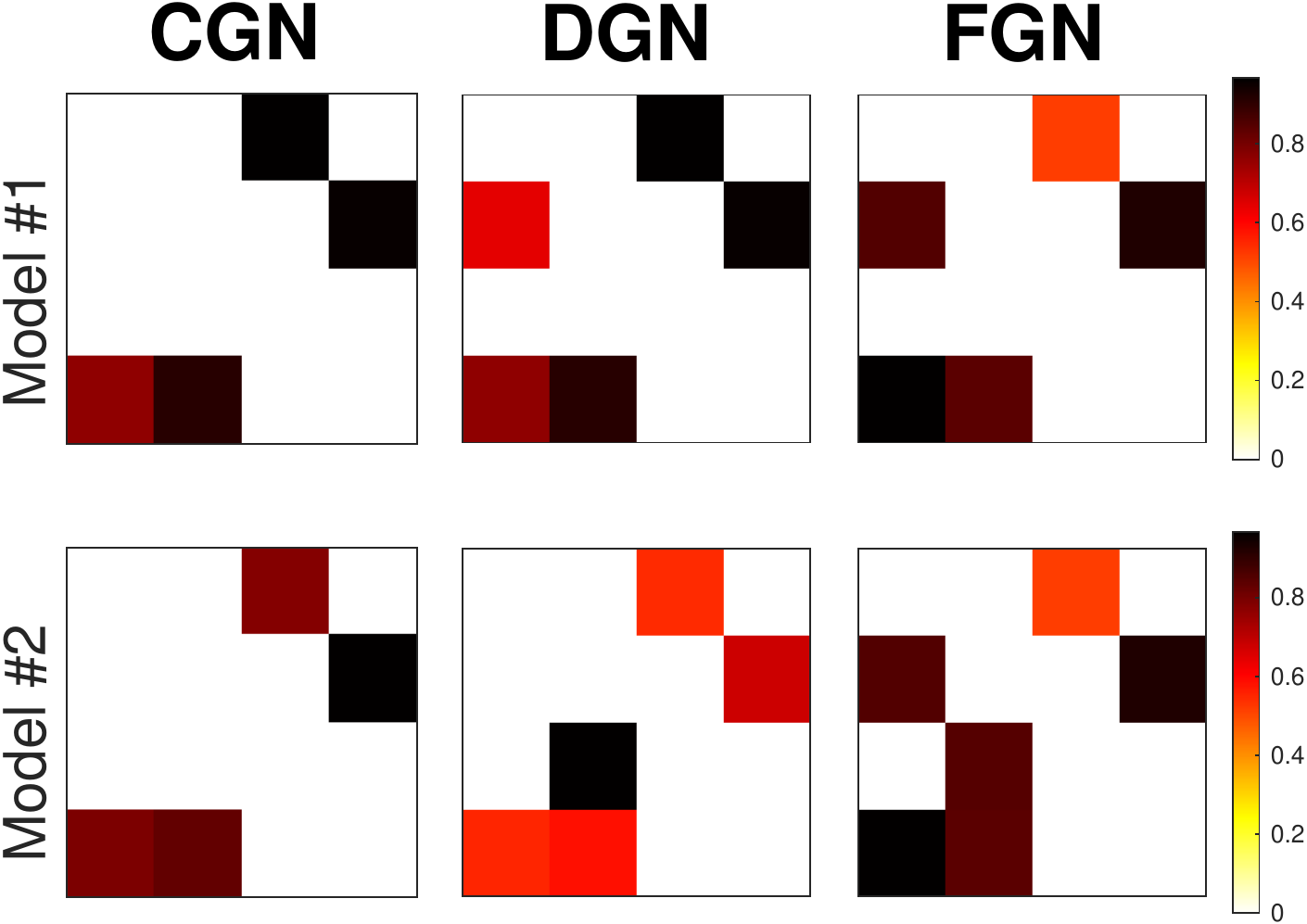}
\caption{Two $n$-dimensional GC networks ($K=2,n=4$) obtained from CGN, DGN and FGN formulations. For each cell, the color map is proportional to $\Vert B_{ij}^{(k)} \Vert $ (darker cells represent stronger GC.)}
\label{fig:graph3GC}
\end{subfigure} 
\begin{subfigure}[c]{0.58\linewidth}
\centering
\includegraphics[width=1\linewidth]{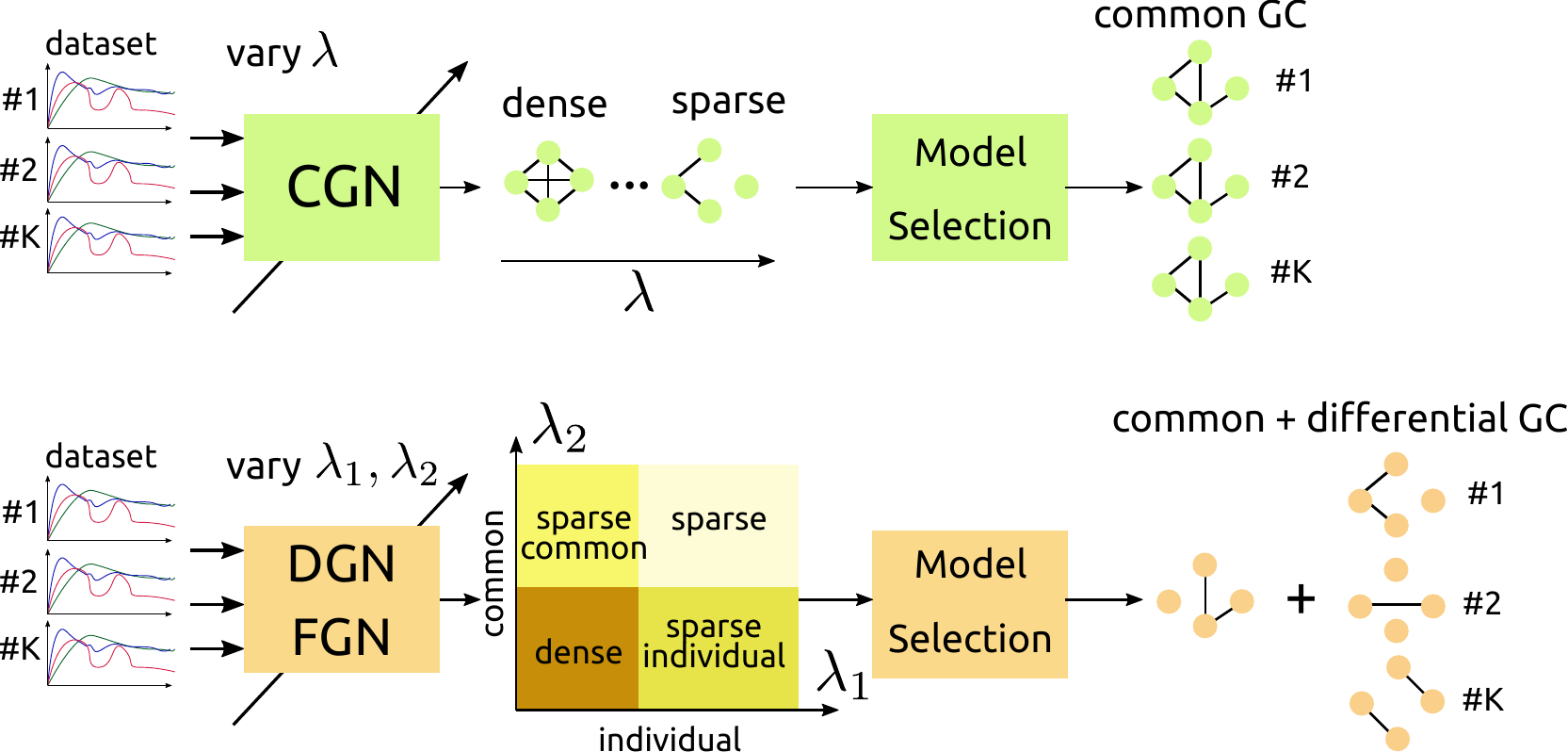}
\caption{Causality learning schemes.}
\label{fig:learning_scheme}
\end{subfigure}
\caption{GC network estimation results from CGN, DGN, and FGN formulations, and a proposed learning scheme.}
\end{figure}

Our causality learning protocol involves extracting one group-level common network and multiple differential networks of individual models as illustrated in \Cref{fig:graph3GC}. The motivation behind the first goal is that, while each model may contain a different intrinsic GC structure, they may also share an underlying meaningful group-level characteristic; CGN's goal is to completely capture these commonalities in its estimated GC networks. The second goal targets the differences among the multiple networks directly, which are captured by DGN and FGN. They first capture the common part by extracting the overlapping nonzeros of $B_{ij}^{(k)}$ among $k$ (\eg, entries $(1,3),(2,4),(4,1),(4,2)$ in \Cref{fig:graph3GC}). The second goal aims at investigating the differences among multiple networks directly, which can be captured by DGN and FGN. After the common part is specified, the differential network of each model correspond to the remaining nonzero locations of $B_{ij}^{(k)}$ for each $k$ (entries $B_{21}^{(1)}$ and $B_{32}^{(2)}$ in \Cref{fig:graph3GC}). 

\Cref{fig:learning_scheme} illustrates our learning scheme for a joint estimation of GC networks. First, we obtain GC networks with various degrees of sparsity by solving the optimization problem \eqref{eq:jointGC} with several penalty parameter values specified. Then, a model selection technique selects the optimal $\lambda$ of CGN or the best pair $(\lambda_1,\lambda_2)$ of DGN/FGN, that minimizes the extended Bayesian Information Criterion (eBIC) score \cite{CC08},
\begin{equation*}
\mathrm{eBIC(\lambda_{1},\lambda_{2})} = -2\mathcal{L}(\lambda_{1},\lambda_{2}) + \mathrm{df} \cdot \log(T-p)+ 2\gamma\log {n^{2}pK \choose \mathrm{df}},
\end{equation*}
where $0\leq \gamma \leq 1$, $\mathrm{df}$ is the degrees of freedom (effective number of model parameters), and the $n$-choose-$k$ term, ${n^{2}pK \choose \mathrm{df}}$, represents the number of possible model candidates with a given $\mathrm{df}$. As an extension of the BIC, the eBIC includes the additional last term which is the log-prior distribution of the model that varies upon the model's $\mathrm{df}$ and a tuning parameter $(\gamma)$ that eventually affects the eBIC's asymptotic property. For CGN, the $\mathrm{df}$ is approximated as the norm ratio of regularized to LS solutions and the number of nonzero groups, following the group lasso framework in~\cite{YL06}. For DGN, $\mathrm{df}$ is the number of nonzero VAR coefficients of all models, while for FGN, the $\mathrm{df}$ is counted once for the fused parameters. The model parameters used for computing the log-likehood, $\mathcal{L}(\lambda_1,\lambda_2)$, are estimated by re-fitting the VAR model subject to the sparsity constraint specified by $(\lambda_1,\lambda_2)$. 

\subsection{Algorithms}
When $q=1$ (convex case) in the penalty functions \eqref{eq:common_penalty}, \eqref{eq:common_diff_penalty}, and \eqref{eq:sim_diff_penalty}, the optimization \eqref{eq:jointGC} is solved using the ADMM (Alternating Direction Method of Multiplier) algorithm with an adaptive penalty ($\rho$) scheme proposed in~\cite{XFG17} to expedite the convergence. For $q=1/2$ (non-convex case), we adapted the technique by~\cite{XLLY17} which heuristically applies the adaptive ADMM with $\rho$ scaled up after a predetermined number of iterations; however, we added a heuristic termination rule that stops the algorithm after the primal residual converges. The global convergence of the ADMM-based algorithm implemented on our non-convex problem is still an open research question; however, if both primal and dual residuals converge, we can guarantee a local optimality of the non-convex problem~\cite{BPC11}. Details of splitting techniques used in ADMM and pseudo codes are described in the \Cref{sec:vector_formulation,sec:alg}.

\section{Simulation results}

\subsection{Experiment setting}
Comparative performances with relevant methods in literature were evaluated on simulations for each formulation. We simulated VAR using~\eqref{eq:jointGC} for $100$ replicates of model parameters with $\Cov(\epsilon^{(k)})=I_{n}$ in various settings. Comparisons illustrated the effects of the ground-truth network density, the number of models ($K$), and the choices of penalty functions. A sparse GC network estimation can be regarded as a binary classification where the positive (negative) class is the nonzero (zero) $B_{ij}^{(k)}$; therefore, typical performance indices -- F1 score, false positive rate (FPR), accuracy (ACC), and Matthews correlation coefficient (MCC) -- were evaluated on the 100 runs. Performance evaluation was also conducted specifically on the: i) \emph{total part}, ii) \emph{common part}, and iii) \emph{differential part} of the multiple GC networks. We have marked our results with \texttt{CGN, DGN} and \texttt{FGN} for those obtained using the non-convex penalty ($q=1/2$), or with an additional \texttt{cvx-} prefix for those of convex penalty ($q=1$). We present F1 and FPR in box plots as the main indicators to compare performances among the different methods and to capture performance variations over data realizations. The averages of all metrics are available in the \Cref{sec:avg_performance}. Note that our reported performances were contributed to by both mathematical properties of formulations and penalty selection using eBIC; the latter is a crucial step as its selection specifies the optimal topology of estimated GC networks. We provide the source codes to all of our experiments in \url{https://github.com/parinthorn/JGranger_ncvx/}.

\subsection{Common GC}
\label{sec:expCGN}
CGN forces all GC networks to have an identical structure. Ground-truth VAR systems were generated with 10\% and 20\% GC density in the common part, and 5\% GC density in the differential part. The different densities of the common part allowed us to evaluate the effects of the ground-truth network density on the CGN performance. The ground truth system parameters were $n=20, p=1, K=5$, and $T=100$; accordingly, we also set $p=1$ in the estimation. We compared the performance indices on the common part of the ground-truth network of both CGN and cvx-CGN to the following formulations. 

\begin{itemize}
\item \textbf{Song17C:} a group lasso approach by~\cite{JSSunpub} (unpublished work) that is a special case of cvx-CGN using $v_{ij} = 1$ in \eqref{eq:common_penalty}.
\item \textbf{Greg15:} a combination of group lasso and Tikhonov regularization approach~\cite{G15}. This work was similar to Song17C but additionally penalized the diagonal of the VAR coefficients with an $\ell_{2}$ penalty. In our opinion, the diagonal is not involved in inferring Granger causality among variables, and thus such regularization only affects the model parameter biases. In \cite{G15}, the penalty parameters of group lasso and Tikhonov regularization were set to be equal.
\end{itemize}

\begin{figure}[h!]
\begin{subfigure}[t]{0.5\linewidth}
    \centering
    \includegraphics[width=\columnwidth]{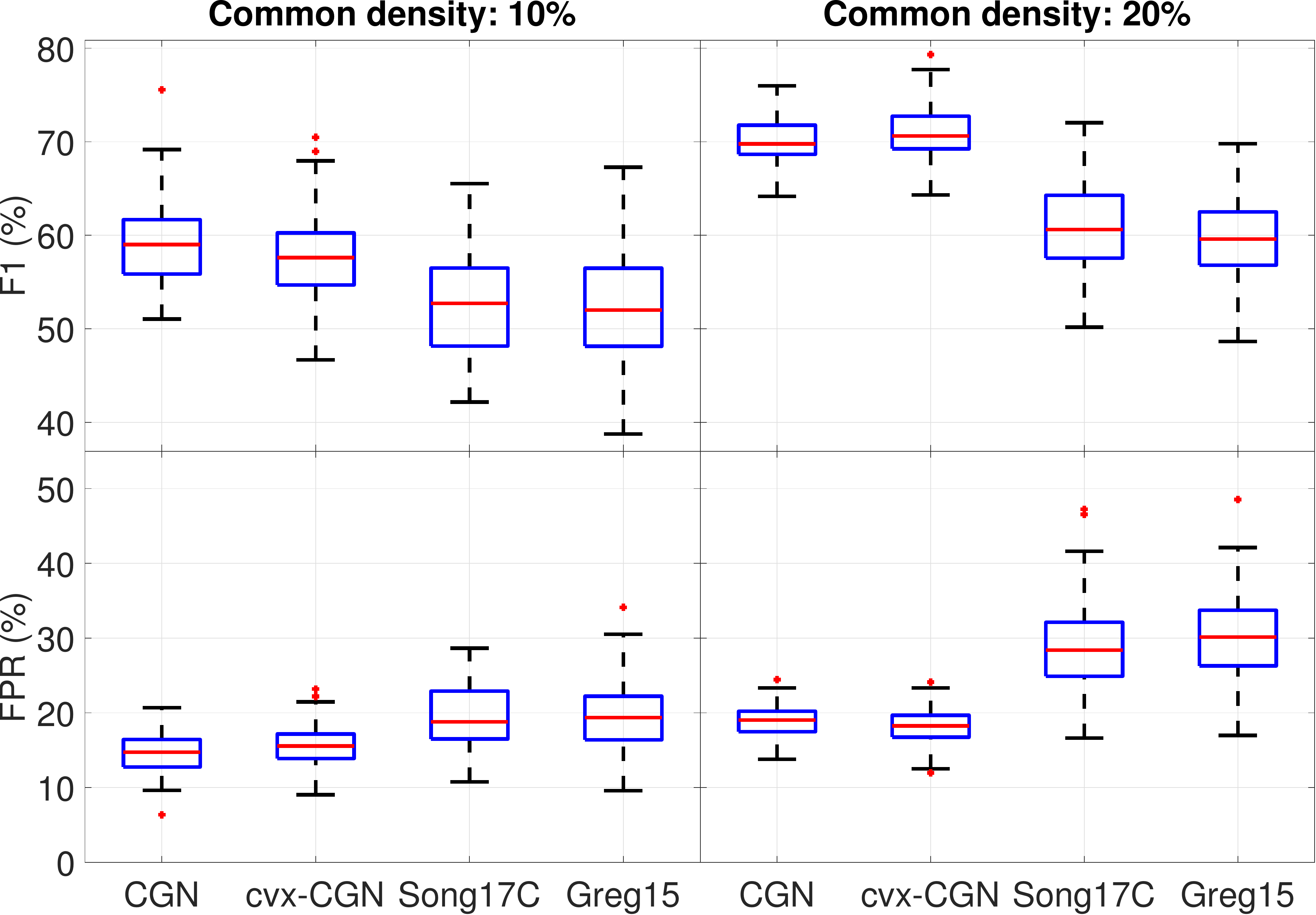}
    \caption{Box plots of F1 and FPR of the estimated common GC. The interquartile ranges of CGN and cvx-CGN were relatively smaller than those of literature methods.}
    \label{fig:exp_commonGC_A}
\end{subfigure} \hfill
\begin{subfigure}[t]{0.48\linewidth}
    \centering
    \includegraphics[width=0.7\columnwidth]{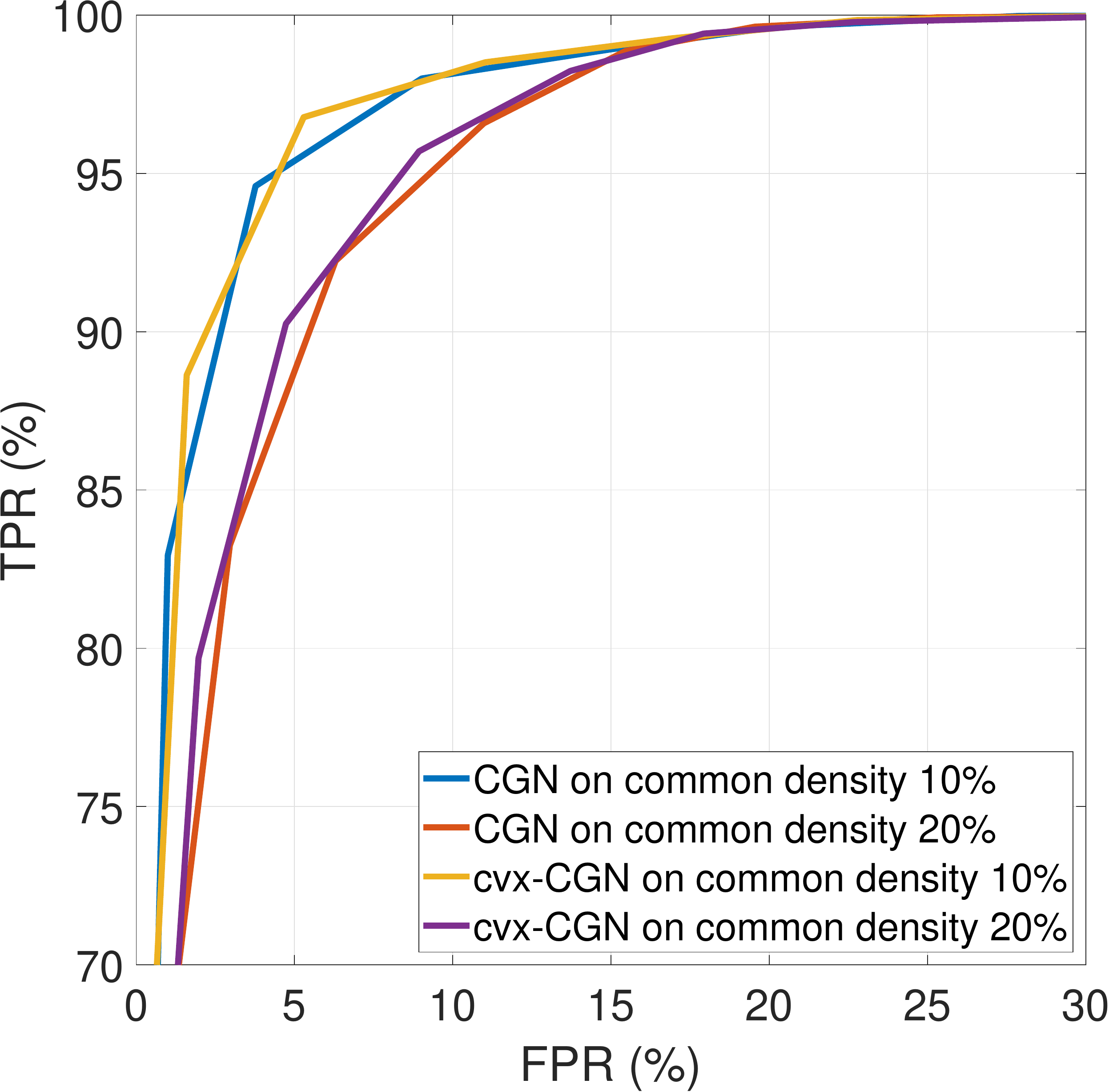}
    \caption{Averaged ROC curve. Each point on ROC corresponds to a value of $\lambda$ in \eqref{eq:common_penalty}. For each method, the area under curve reduces as the density increases.}
    \label{fig:exp_commonGC_B}
\end{subfigure}
\caption{Performances of \textbf{CGN} in estimating ground-truth common GC networks of $\mathbf{10}\%$ and $\mathbf{20}\%$ density.}
\label{fig:cgn}
\end{figure}

The results in \Cref{fig:exp_commonGC_A} show that both CGN and cvx-CGN outperformed the methods in literature, with significant improvements in the case of 20\% common density. This directly came from introducing a reasonable choice of $v_{ij}$ to incorporate the different likelihood of zero locations in VAR parameters, unlike setting equal $v_{ij}$'s in~\cite{JSSunpub}. Meanwhile, the Tikhonov regularization of Greg15 had no direct effect on the GC estimation's performance as the evaluation was conceptually taken only on the off-diagonal VAR coefficients. The effects of common density on our formulations are illustrated in \Cref{fig:exp_commonGC_B}, where the performances dropped as the density increased, which is typical of sparse-inducing frameworks. However, the effect on the actual performance that is affected by the penalty selection (\Cref{fig:exp_commonGC_A}) showed the opposite, where the F1 score increased alongside the density.  

\subsection{Common and differential GC}
\label{sec:expDGN}
The DGN formulation is intended to estimate both the common and differential parts of GC networks. To test this, we generated ground-truth models with a common part of 10\% GC density and differential parts with various densities of 1\% and 5\%. Model parameters were set as $n=20, p=1, K=5$, and $T=100$, and we selected $p=1$ in the estimation. We also examined varying $K=5$ and $K=50$ while setting the differential density to 5\% to determine how the number of models impacts the DGN's performance. Other methods in the literature which also decompose GC networks into common and differential parts for estimation are:

\begin{figure}
\begin{subfigure}[t]{0.48\linewidth}
\centering
\includegraphics[width=\columnwidth]{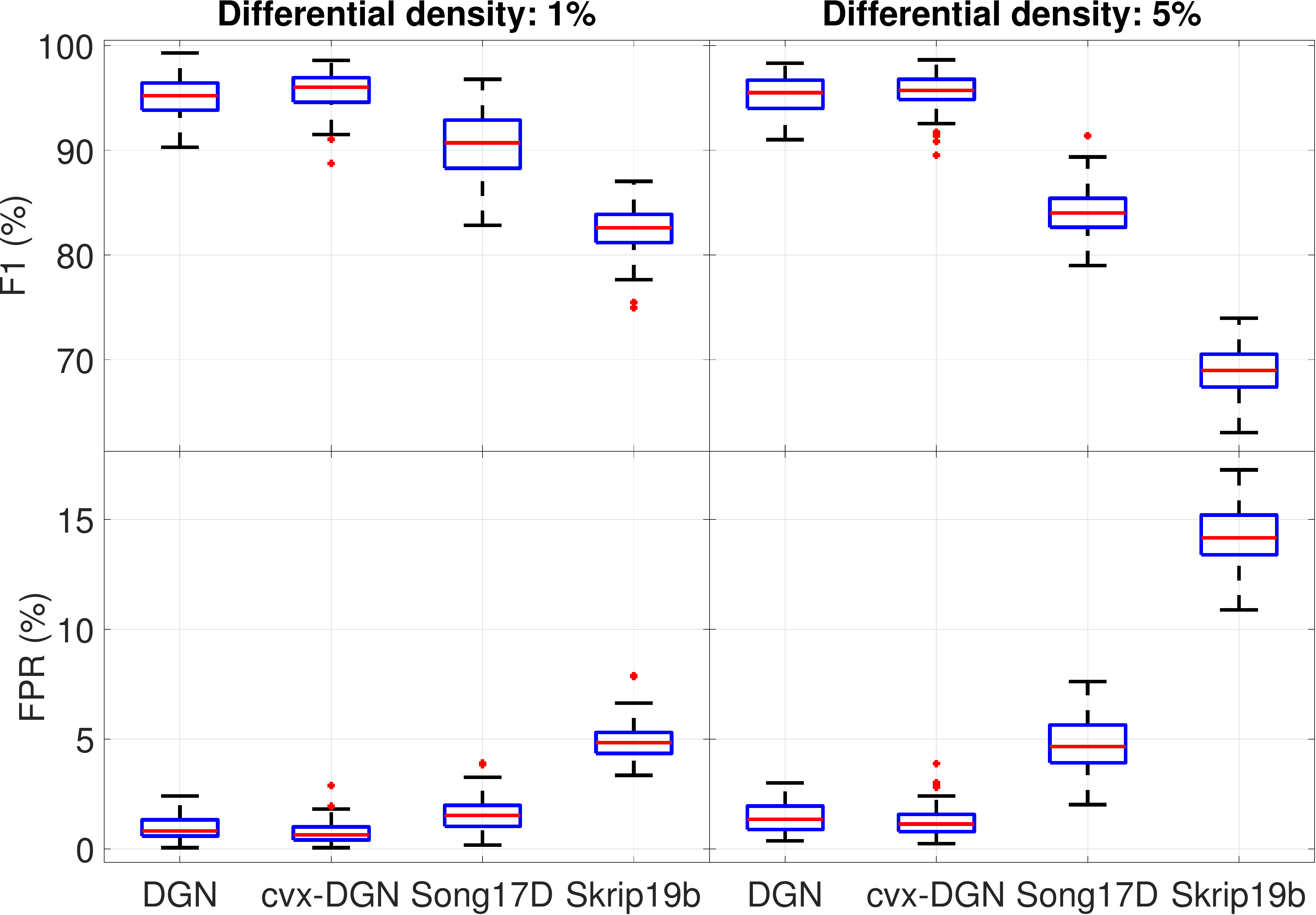}
\caption{Vary the densities of ground-truth differential GC.}
\label{fig:exp_diffGC_A}
\end{subfigure} \hfill
\begin{subfigure}[t]{0.48\linewidth}
\centering
\includegraphics[width=\columnwidth]{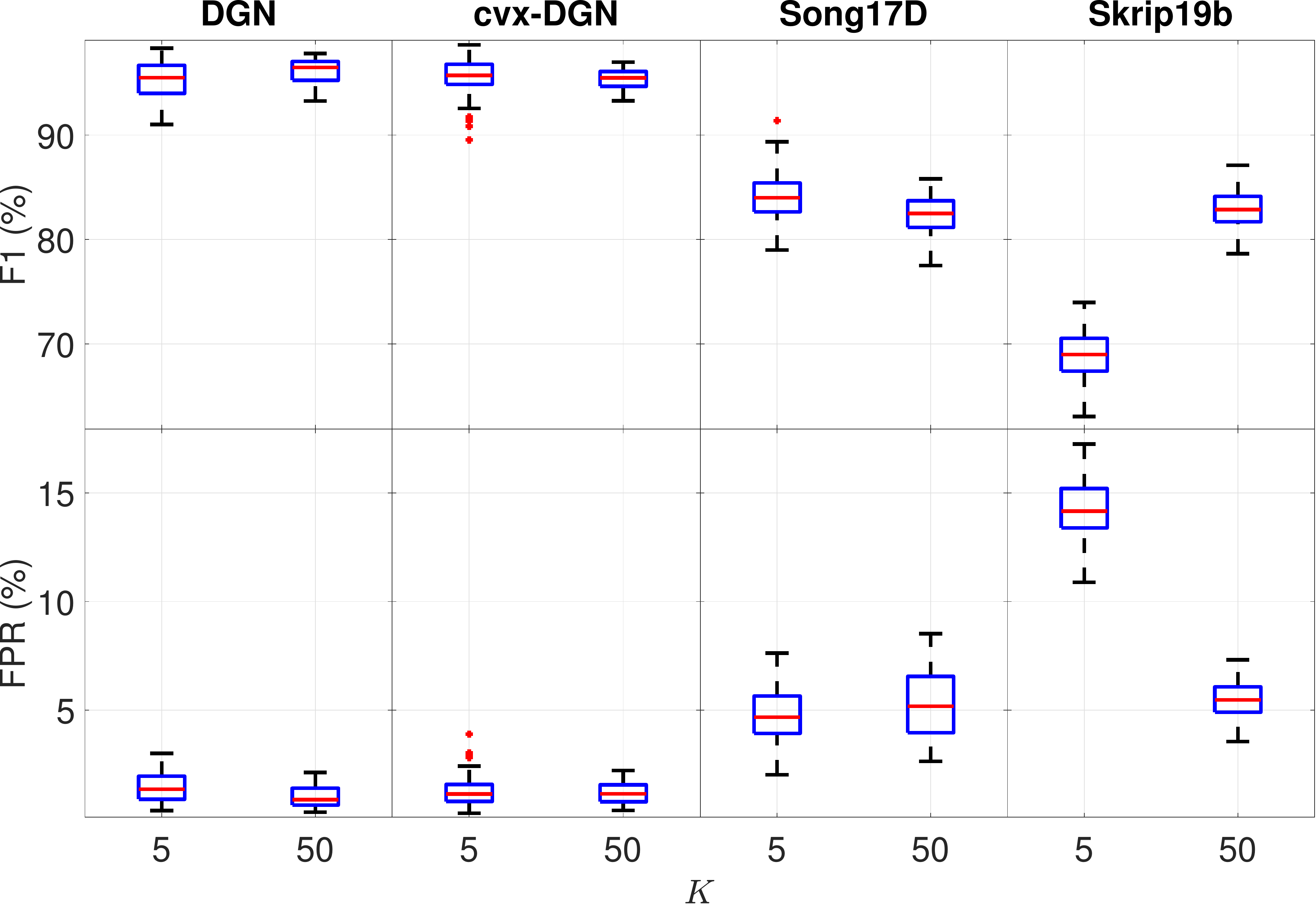}
\caption{Vary the number of models ($K$).}
\label{fig:exp_diffGC_B}
\end{subfigure}
\caption{Performances of \textbf{DGN} and other methods. The interquartile ranges of DGN and cvx-DGN were relatively smaller than those of literature methods.}
\label{fig:dgn}
\end{figure}

\begin{itemize}
\item \textbf{Skrip19b}: a two-stage approach~\cite{SkM:19b} that first estimated the parameters of the common network using a group lasso (like our CGN) and then estimated the individual components based on the resulting common network. This approach does not guarantee a global optimal solution as the parameters were estimated in sequential steps, not being optimized in batch. The number of models was found to impact this approach's performance, with results improving as the number of models increased. 
\item \textbf{Song17D}: a group lasso combination approach~\cite{JSSunpub} which is essentially the same as the cvx-DGN but with penalty weights set to unity instead of being specified as in this experiment. 
\end{itemize}

In \Cref{fig:exp_diffGC_A}, as the differential density varied, our performances were desirably less sensitive to this change than both Song17D and Skrip19b. Since Song17D does not have any priors, its poorer performance is expected as compared to cvx-CGN. Meanwhile, Skrip19b performance may suffer from sub-optimal penalty selection since penalties were sequentially deduced and thus depended on one another. 

In \Cref{fig:exp_diffGC_B}, as $K$ increased, Skrip19b's performance improved as previously claimed. While DGN and cvx-DGN's performances appeared independent of $K$, they both outperformed Song17D and Skrip19b for both $K=5$ and $K=50$. For now, please note that the number of models should directly affect computational complexity due to the different number of variables; later in the Limitation Section, we will discuss in detail the effects of
$K$ on the \emph{separate} performances of common and differential network estimation.

\subsection{Fused and differential GC}
\label{sec:expFGN}

To evaluate FGN's performance, ground-truth systems similar to that described in \Cref{sec:expDGN} were generated except with the common-part VAR parameters restricted to being all equal to examine if the fused framework can identify the locations of this common ground. FGN's performances were compared with the following methods:
\begin{itemize}
\item \textbf{Skrip19a}: a sparse fused-lasso approach~\cite{SkM:19a} which employed a combination of lasso and fused lasso to induce a sparsity on VAR coefficients and model parameter differences, respectively. The VAR sparsity obtained from the lasso term does not correspond to the characterization of GC on the all-lag VAR coefficients (as discussed in \Cref{sec:fgn}). While their framework can be applied for any VAR order ($p$), the available source codes can only be implemented with $p=1$.
\item \textbf{Song15}: a group fused-lasso approach~\cite{Son:15} similar to cvx-FGN, except that the fused term was only taken on the consecutive models and the relative penalty weight was set to one (as described in \Cref{sec:fgn}).
\end{itemize}

\begin{figure}
\begin{subfigure}[b]{0.48\linewidth}
\centering
\includegraphics[width=\columnwidth]{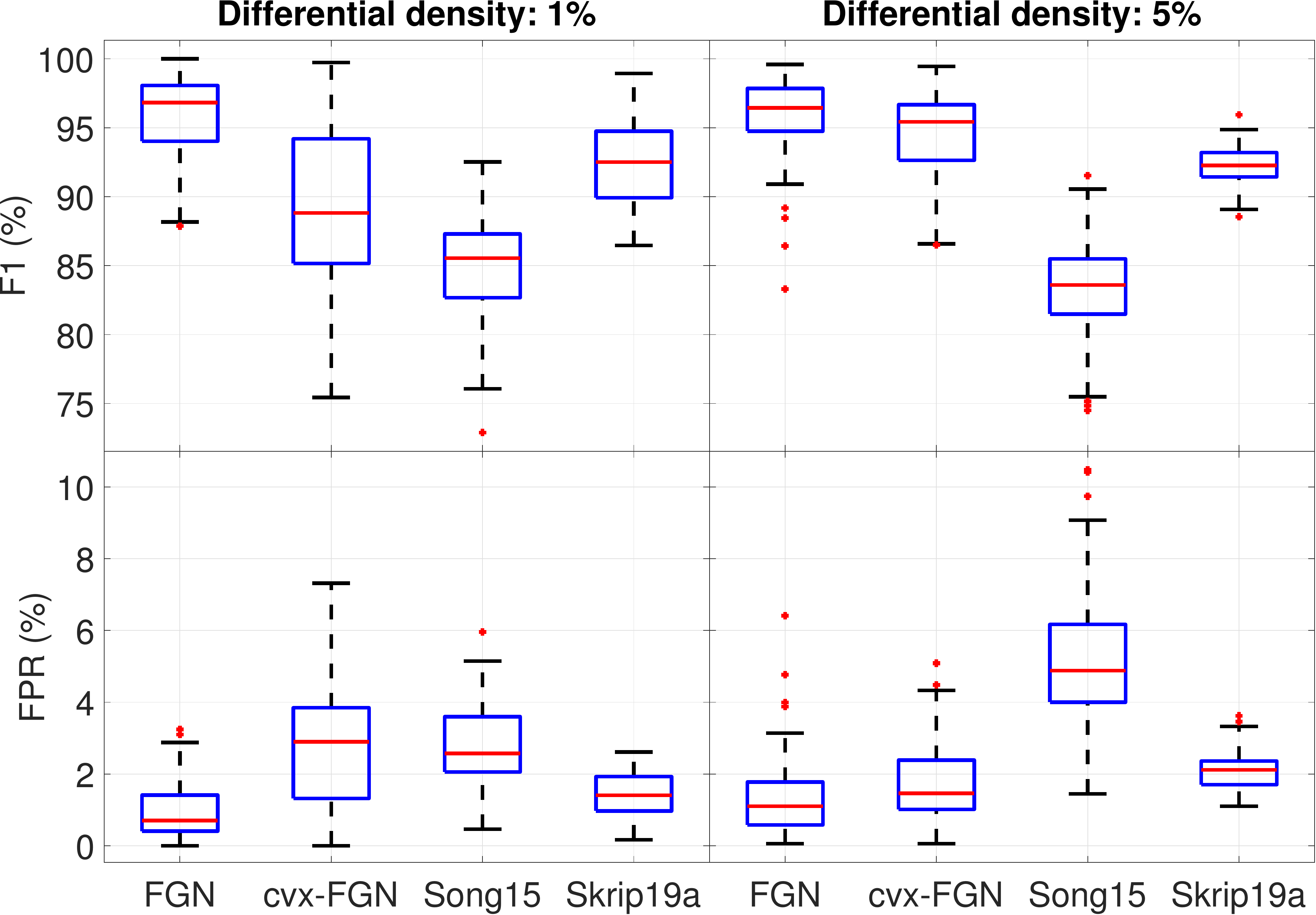}
\caption{Performances of GC network estimation.}
\label{fig:exp_fusedGC}
\end{subfigure} \hfill
\begin{subfigure}[b]{0.48\linewidth}
    \centering
    \includegraphics[width=\columnwidth]{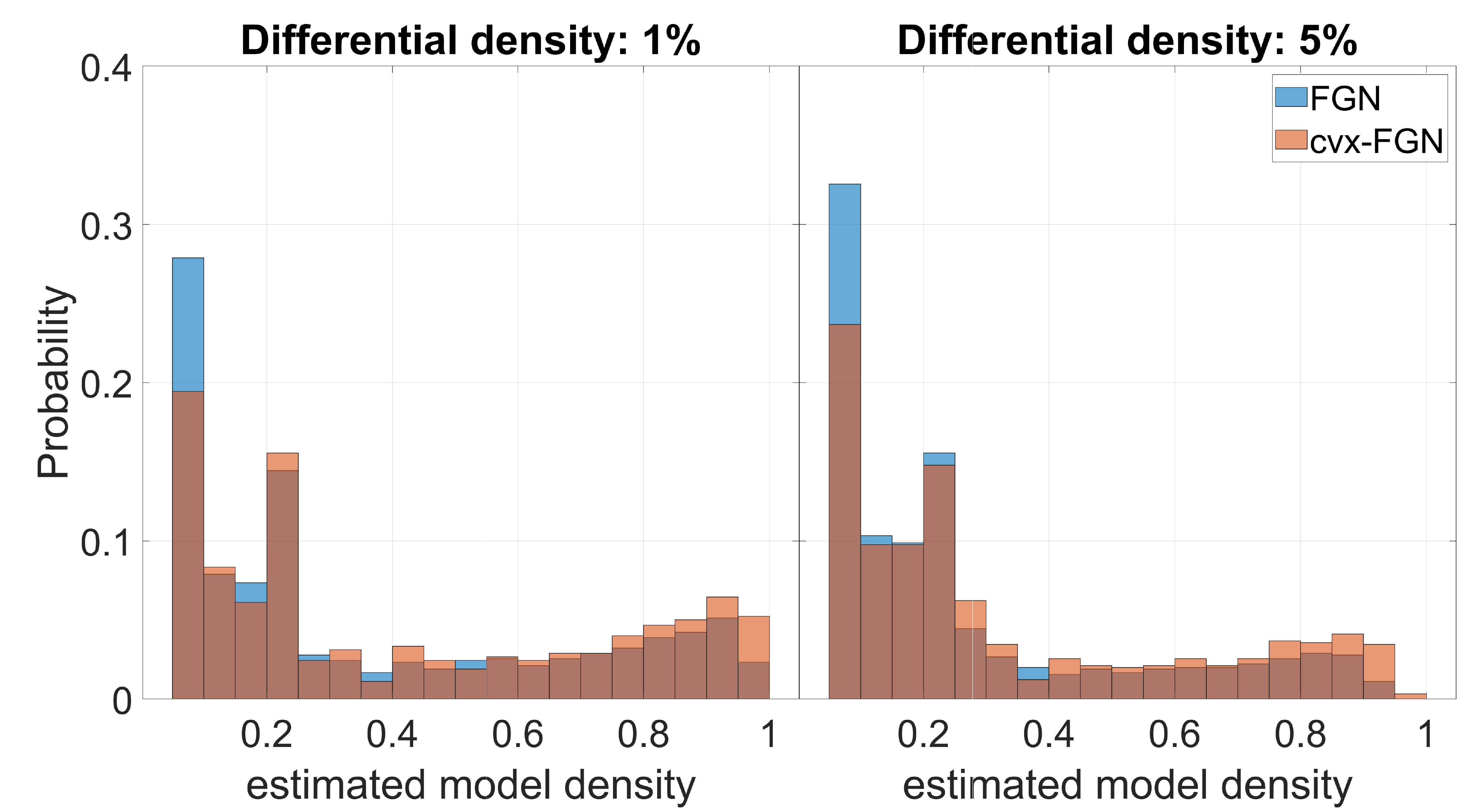}
    \caption{Histograms of the model density (counted as the estimated model's degree of freedom, scaled to 1) under the two cases of the ground-truth differential density.}
    \label{fig:exp_FGN_density_histogram}
\end{subfigure}
\caption{Performances of \textbf{FGN} as the density of ground-truth differential GC networks varied.}
\label{fig:fgn}
\end{figure}

\Cref{fig:exp_fusedGC} shows that FGN outperformed the other methods in all settings, thanks to its formulation that accommodates a prior specifying identical parameters across models. Also, FGN appeared to be more robust to variations in GC density than the other methods. We can see that as GC density increased, the performances of Song15 and Skrip19a deteriorated, as is generally expected in sparse learning; however, cvx-FGN's performance unexpectedly increased alongside differential density. This can be explained by the histogram in \Cref{fig:exp_FGN_density_histogram} that shows the empirical distributions of the estimated model's degree of freedom as $(\lambda_1,\lambda_2)$ varied. For cvx-FGN, the portion of extracted sparse models in the 1\%-density setting was less than at 5\%, leading to a smaller number of sparse model candidates for eBIC to choose, and hence, less likelihood to obtain a high-performing estimated model that is also sparse. Comparing between FGN and cvx-FGN, the former’s non-convex penalty resulted in a higher portion of extracted sparse models, which provides eBIC more and better sparse candidates to choose from.

\subsection{Advantage of non-convex penalties}
\label{sec:expNCVX}
\begin{figure}
    \centering
    \includegraphics[width=0.5\linewidth]{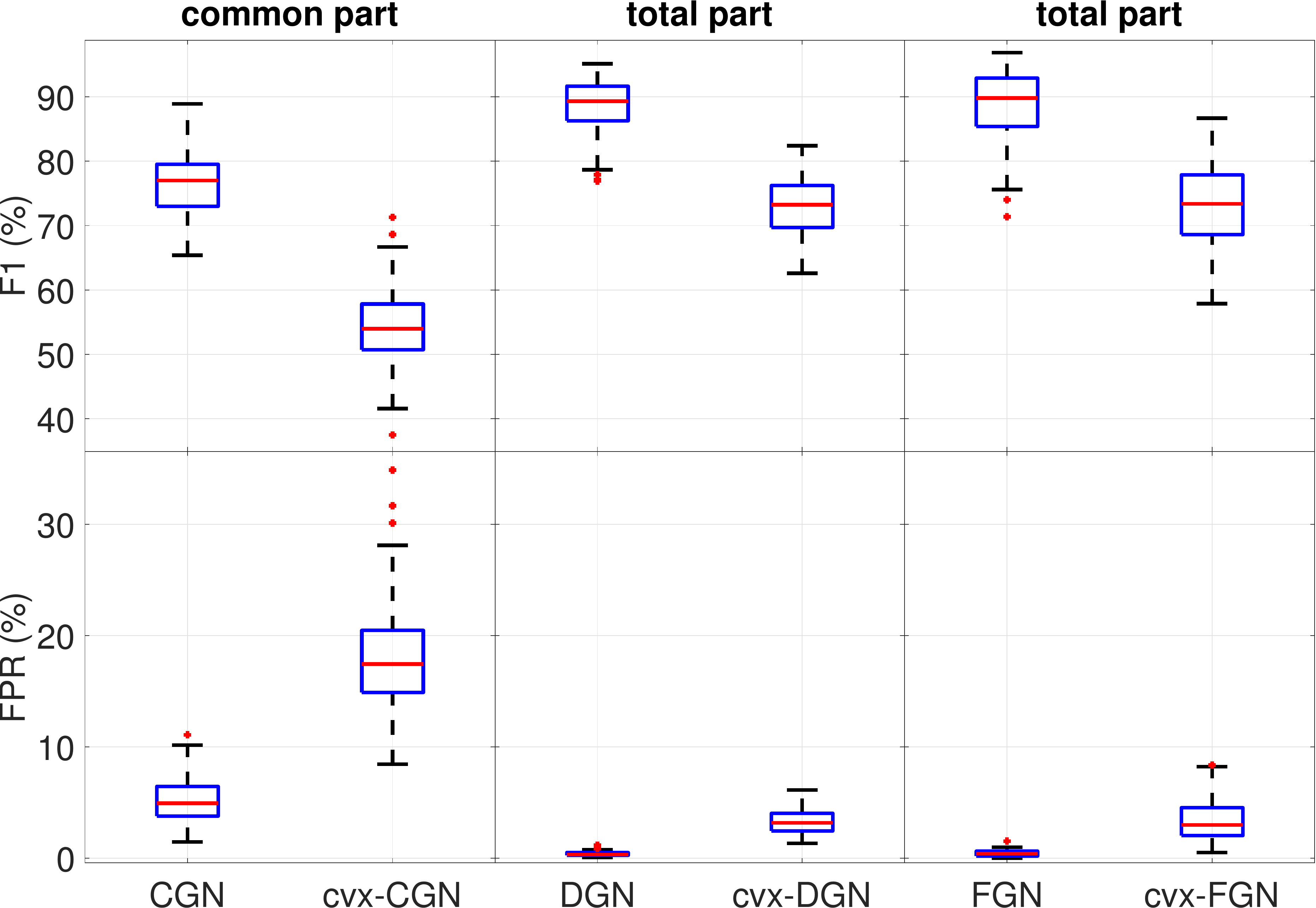}
    \caption{Performance comparison of CGN, DGN, FGN and their convex penalty counterparts. The evaluation of CGN and cvx-CGN was performed on the common GC part.}
    \label{fig:ncvx_advantage}
\end{figure}

The estimation error bound presented in \cite{HLMQY17} depends on the ground-truth group-sparsity level. The results in \Cref{fig:cgn,fig:dgn,fig:fgn} did not show significant differences between the convex and non-convex performances; perhaps this was due to the ground-truth systems being relatively sparse and possibly within true sparsity levels that are in-range for the two formulations to perform closely. Also, the GREC assumption used by \cite{HLMQY17} to obtain a recovery bound is prone to be violated in a low-sample-high-dimension setting. As such, in order to illustrate the benefits of the non-convex penalty over the convex penalty, we increased the ratio of variables to data samples from 4:1 to 8:1 by setting the ground-truth system parameters as $(n,p,K,T) = (20,1,5,100)$ and $(20,3,5,150)$. Also, the densities of the common and differential GC were set to 10\% and 5\%, respectively. The results, seen in \Cref{fig:ncvx_advantage}, showed the non-convex formulations outperforming their convex counterparts. Under a fixed penalty parameter, the non-convex formulations yielded sparser solutions than the convex ones. Also, the non-convex formulations generated lower false positive rates than the convex formulations when the true system was sufficiently sparse. However, we note that the benefit of non-convex formulations should be carefully deemed with their limitations from algorithm point of views, as to be later discussed in \Cref{sec:limitation}.

\section{Application to fMRI data}

Adolescents with attention deficit hyperactivity disorder (ADHD) suffer from abnormalities of some brain regions, both functionally and structurally, when compared to typically developing children (TDC). We tried to identify the differences of effective brain connectivity underlying the two groups using the ADHD-200 competition data set~\cite{BCC17}. We selected 18 age-matched male subjects of 7-17 years old from each of the ADHD and TDC groups. The resting-state fMRI time series were averaged over voxels within the AAL-atlas regions of interest (ROIs), resulting in 116-channel time series, each with 172 time points. More details of data selection and preprocessing are described in the \Cref{sec:fmridata}.

\subsection{Experiment setting} 
\begin{figure}[h!]
\centering
\includegraphics[width=0.8\linewidth]{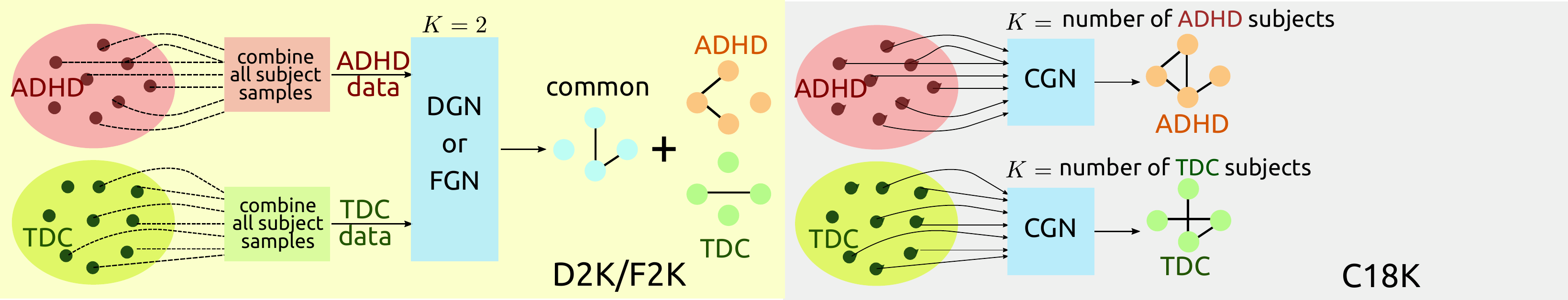}
\caption{Three schemes of learning ADHD and TDC networks.}
\label{fig:adhd_scheme}
\end{figure}

Three schemes for learning the ADHD and TDC networks are presented in \Cref{fig:adhd_scheme}. 
\begin{enumerate}
    \item \textbf{D2K:} For each of TDC and ADHD groups, we pooled data from all subjects, so the sample sizes of each group increased. Each group's combined data served as the input to cvx-DGN\footnote{As the number of samples is sufficiently moderate compared to the number of variables, the distinction between the non-convex and convex formulations may not be significant. For this reason, we employed only the convex formulations to avoid the local optimum, or the algorithm convergence issues.} with $K=2$. The outputs were ADHD and TDC networks with distinguished common and differential parts.
    \item \textbf{F2K:} The scheme was similar to D2K but with cvx-FGN in place of cvx-FGN.
    \item \textbf{C18K:} This scheme employed cvx-CGN to learn a common network among the 18 subjects ($K=18$) in each of the ADHD and TDC data sets. 
\end{enumerate}

Each scheme resulted in two estimated GC networks, one each for the ADHD and TDC subjects. The networks obtained by D2K/F2K contained a common structure across the ADHD and TDC groups, but also a different structure that explained the individual characteristics of each group. Meanwhile, the networks obtained by C2K showed the common dominant connections across subjects in each group. With the number of AAL ROIs at 116, the results are too complicated to visualize as a graphical model. Therefore, we analyzed the brain connections in the estimated GC network using the \emph{edge betweenness centrality} measure~\cite{RS10} used in network theory. This score reflects the number of times that an edge of interest appears in all existing shortest paths between any two nodes of a graph. To compute this score, the weights of all edges in the graph are required. Since higher GC values indicate a stronger connection between two brain regions, we used the reciprocal of GC, or specifically, $1/\Vert B^{(k)}_{ij} \Vert_2$ as a proxy-distance between two regions $i$ and $j$, such that a stronger GC connection would result in a shorter path. If the difference of edge centrality between the ADHD and TDC networks is significantly high (\ie, loosely speaking, the brain connection of the ADHD network is fundamentally different from the TDC), then the percentage that such a GC edge in the ADHD network appears in the shortest paths of the graph would be significantly different from the TDC network. We divided the centrality differences into two types; centrality measures which were lower in ADHD subjects were denoted as the \emph{missing} type, whereas those that were higher in ADHD subjects were denoted as the \emph{extra} type. Brain connections corresponding to the three highest centrality differences are presented in~\Cref{table:ADHD}.

\subsection{Brain connectivity results}

\definecolor{Lightyellow}{rgb}{0.9,1,0.5}
\definecolor{LightCyan}{rgb}{0.88,1,1}

\begin{figure}[h!]
\centering
\begin{subfigure}[b]{0.25\linewidth}
    \centering
    \includegraphics[width=1\linewidth]{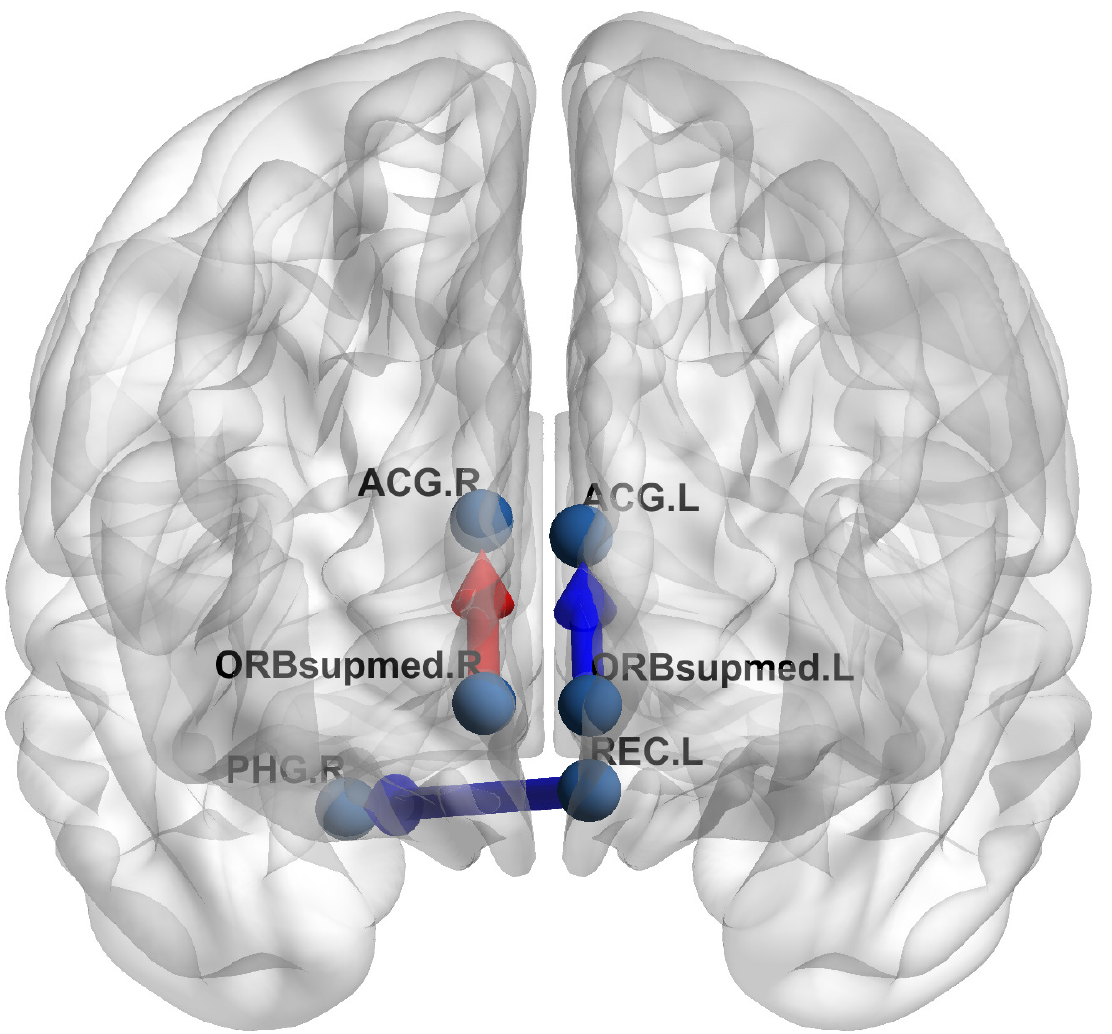}
    \caption{Coronal view}
    \label{fig:orbitlimbic_front}
\end{subfigure} \qquad
\begin{subfigure}[b]{0.25\linewidth}
    \centering
    \includegraphics[width=1\linewidth]{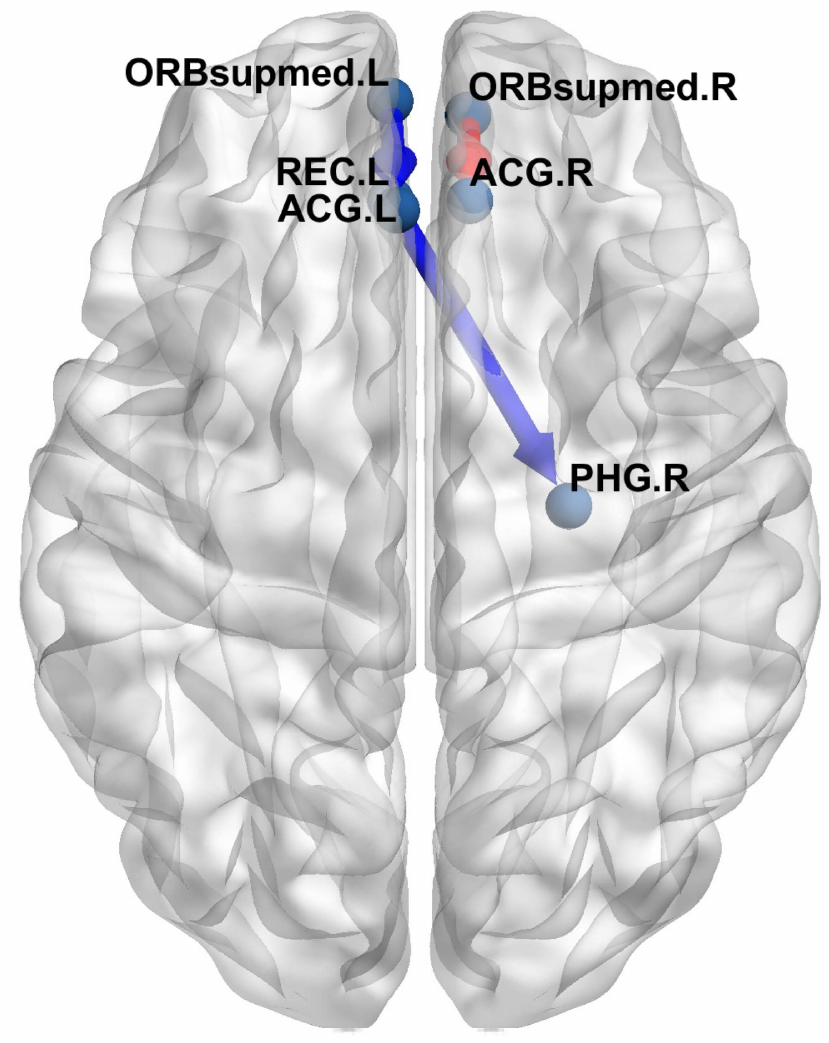}
    \caption{Axial view}
    \label{fig:orbitlimbic_top}
\end{subfigure}
\caption{Selected brain connections among the orbitofrontal region and part of limbic system that have distinct centrality differences between the ADHD and TDC networks. The \textcolor{blue} {blue} (\textcolor{red}{red}) directed edges show a \textcolor{blue}{missing} (\textcolor{red}{extra}) connections in ADHD.}
\label{fig:orbitlimbic}
\end{figure}

\begin{table}[t]
\centering
\caption{Brain connections that indicate connectivity differences between ADHD and TDC, ranked by the three highest absolute differences of the edge betweenness centrality.}
\label{table:ADHD}
\resizebox{1\linewidth}{!}{
\begin{tabular}{|c|l|p{6.5cm}|p{6.5cm}|>{\centering}m{1.7cm}|l|} \hline
\bf No. & \bf Scheme & \bf Cause & \bf Effect & \bf Centrality difference & \bf Associated system \\ \hline
\rowcolor{Lightyellow}
\multicolumn{6}{|c|}{ADHD $<$ TDC (missing)} \\ \hline
1 & D2K&Anterior cingulate gyrus L&Anterior cingulate gyrus R&-648&Limbic system \\ 
2 & &Anterior cingulate gyrus R&Fusiform gyrus R&-474&Limbic-Temporal \\ 
3 &&Cerebellum 3 L&Anterior cingulate gyrus L&-442&Cerebellar-Limbic \\ 
4 & F2K &Rectus gyrus L&Parahippocampal gyrus R&-282&Orbitofrontal-Limbic \\
5 & &Amygdala L&Superior temporal gyrus L&-141&Limbic-Temporal \\
6 & &Parahippocampal gyrus R&Inferior frontal gyrus (orbital) R&-130&Limbic-Frontal \\
7 & C18K&Superior frontal gyrus (medial orbital) L&Anterior cingulate gyrus L&-349&Orbitofrontal-Limbic \\
8 & &Superior frontal gyrus (medial orbital) L&Superior frontal gyrus (medial orbital)  R&-125&Orbitofrontal \\
9 & &Rolandic operculum R&Precentral gyrus R&-124& \\  \hline
\rowcolor{LightCyan}
\multicolumn{6}{|c|}{ADHD $>$ TDC (extra)} \\ \hline 
10 & D2K&Superior frontal gyrus (medial orbital) R&Anterior cingulate gyrus R&367&Orbitofrontal-Limbic \\
11 & &Olfactory cortex L&Insula L&344&Olfactory-insular \\
12 & &Temporal pole (superior) L&Putamen L&327&Temporal-Frontal \\
13 & F2K&Middle frontal gyrus (orbital) R&Superior frontal gyrus (orbital) R&164&Orbitofrontal \\
14 & &Temporal pole (superior) L&Rolandic operculum L&164&Temporal-Operculum \\
15 & &Rectus gyrus R&Superior frontal gyrus (medial orbital) R&143&Orbitofrontal \\
16 & C18K&Supplementary motor area R&Precuneus R&306& \\
17 & &Superior frontal gyrus (medial orbital) L&Middle frontal gyrus (orbital) L&296&Orbitofrontal \\
18 & &Middle frontal gyrus (orbital) L&Superior frontal gyrus (medial orbital) R&282&Orbitofrontal \\ \hline
\multicolumn{6}{l}{* L/R denotes the left or right hemisphere. ** The regions with (orbital) or (medial orbital) are orbitofrontal area. } \\ 
\multicolumn{6}{l}{*** All the missing-type links were not in the ADHD network, except no.3. All the extra-type links were not in the TDC network, except no.18.} \\ 
\end{tabular} 
}
\end{table}
\Cref{table:ADHD} shows that the differences between the ADHD and TDC subjects' brain structures were primarily concentrated in the \emph{orbitofrontal region (ORB)}, and regions associated with the \emph{limbic system}. The orbitofrontal region is associated with a reward-motivation system that responds to rewards or punishments~\cite{RCF20}. As for the limbic system, two ROIs identified in this study were i) the anterior cingulate cortex (ACG), which was related to emotion~\cite{BLP00}, decision making, and social interaction \cite{LMM13}, and ii) the parahippocampal gyrus (PHG) that was involved in memory retrieval and emotion processing~\cite{AKB13}. Among these regions, the ADHD and TDC brain structures showed significantly different centrality scores in three brain connections, as shown in \Cref{fig:orbitlimbic}.

For the first difference, link no.7 connecting from \emph{the left superior frontal gyrus (ORBsupmed) to the left ACG} was missing from ADHD subjects' brain structures. Anatomical evidence has shown that ADHD subjects with focal brain damages in these two areas exhibit anti-social behavior \cite{Bech04}; this was later supported by the discovery that decreased functional brain connectivity between these two regions was correlated to subjects exhibiting social anxiety disorder \cite{HSW11}. The second difference between ADHD and TDC subjects, however, was that the ADHD subjects possessed an \emph{extra-type} link no.10 in their \emph{right brain hemispheres}, which actually connects the same analogous regions of the left hemisphere as link no.7; see the coronal view in \Cref{fig:orbitlimbic_front}. Previously, \cite{TV12} had found a similar higher functional connectivity between the ORB and ACG regions of ADHD subjects’ brains. By replicating these previously confirmed results, we are convinced that the increased centrality score in the right hemisphere of the ADHD network accounts for a reward-motivation dysfunction, and the decreased score in the left hemisphere may explain anti-social behavior. Thirdly, connection no.4 from the \emph{left rectus gyrus (REC) to the right PHG} was missing in the ADHD network, indicating a broken connection from the ORB region to part of the limbic system. This was also partly supported by \cite{IRL19} which used decision trees to discover that features extracted from the PHG region was highly discriminative for ADHD classification. In addition, \cite{SST16} also discovered that subjects with REC resection (similar to our missing connection no.4 link) had memory recall and language skill impairments when tested with the mini-mental state examination, thus agreeing with our results that the missing REC may involve with limbic functions. 

Also listed in \Cref{table:ADHD} are five connections \emph{within} the orbitofrontal system (no.8,13,15,17,18) that had distinctive centrality scores between ADHD and TDC subjects. For ADHD patients, the ORB region was responsible for reward learning sensitivities or a slower learning rate when the objective of a reward-related task was changed \cite{IU02}. Out of the five connections, no.8 and 17 shared a common origin of the left ORBsupmed, a region whose features had previously been extracted and effectively used for ADHD classification \cite{IRL19}. Elsewhere, the extra connection (no.14) from \emph{the right REC to the right ORBsupmed} in ADHD subjects was previously found to be significant for ADHD classification using linear discriminant analysis by \cite{TLC20}. Other reports have found that the REC and ORBsupmed regions were responsible for reward-system dysfunctions in a study of subjects with binge eating disorder \cite{SCM14}, symptoms that were related to ADHD as reported in \cite{SRB16}.

\subsection{Choices of formulation}
We can draw two conclusions regarding how the three formulations infer different brain connections with the results in \Cref{table:ADHD}. First, most links discovered by C18K were concentrated within the ORB region, whereas D2K/F2K discovered connections from a more diverse group of ROIs. In C18K setting, the ADHD and TDC networks were estimated separately, and each network revealed significant common characteristics within each group. So while the ORB can separately explain brain functionalities for either the ADHD or TDC groups, the discovered centrality and ORB subregion differences were also concentrated within this region. On the contrary, the F2K/D2K formulations estimated the ADHD and TDC networks allowing for network differences to present freely in any regions, and therefore the network differences between the two groups occurred in several areas.

Second, despite using the same paradigm in the F2K and D2K settings, the magnitudes of centrality differences from D2K were higher than those from F2K. This arises from the penalty being used, since the similar parameters across models affects the weight of GC networks. F2K used the fused lasso to encourage parameter similarity between the ADHD and TDC models, resulting in small centrality differences between the two groups. In contrast, D2K used the group lasso to enforce a differential structure in each individual network separately, allowing individual model parameters to shrink by different degrees, and therefore resulting in larger centrality differences. This conclusion suggests that the centrality difference ranking should be made on each formulation separately since the scores are on a different scale.

\section{Limitation and recommendation}
\label{sec:limitation}

\paragraph{Varying penalty parameters.} The causality learning scheme presented in \Cref{fig:learning_scheme} employed the proposed formulations to estimate models with various degrees of sparsity through adjusting $(\lambda_1,\lambda_2)$. For sparse formulations that have a single penalty parameter (such as lasso, group lasso, or fused lasso), it is possible to derive a range of the penalty parameters in closed-form, ordered by the model sparsity they induce from densest to sparsest. This range generally depends on the sample size and problem data. Unfortunately, for sparse-inducing problems with two or more penalties, it is difficult to derive such a range analytically. Due to this limitation, a heuristic approach is needed to create a range for $(\lambda_1,\lambda_2)$ by setting the upper bound of $\lambda_1$ to its critical value (as if there was only $\lambda_1$ in the formulation) and vary $\lambda_2$ until the solution is zero. 

\paragraph{Algorithms of non-convex formulation.} Convergence to a global optimum for non-convex problems generally depends on the algorithm initialization. As also pointed out in \cite{WCLQ18}, non-convex penalties may not show any improvements or even distinctions over convex penalties for some choices of initialization (such as zero in the regression problems.) In our implementation, the algorithm of non-convex formulations started with the least-squares solution. When solving the problem with a series of $(\lambda_1,\lambda_2)$, a common remedy is to use the solution associated with the previous pair to initiate the algorithm.

\paragraph{Overlapped penalization.} DGN's penalty consists of two terms that penalize some overlapping groups of parameters. As such, the estimated common part, $C_{ij}$, is also affected by the regularization of the differential part, $B^{(k)}_{ij}$. For a large $K$, as we varied the pair $(\lambda_1,\lambda_2)$ on a grid range, the best solution (in terms of highest F1) evaluated on the common network can be much different from the one evaluated on the differential network; see more experimental results in the \Cref{sec:overlapped_dgn}. In other words, the separate best-case performances of the differential and common networks cannot co-exist using the same pair of $(\lambda_1,\lambda_2)$ for large $K$. The choice of relative weights, $v_{ij}$ and $w^{(k)}_{ij}$, partly mitigates this issue but it does not completely solve the problem. However, achieving the best performance on both the common and differential parts at the same time may not be necessary in practice, since we generally focus either on the common or the differential GC when analyzing results. In settings where the common GCs are more informative, we can select a model that benefits evaluating the common sparsity pattern, and not worry about the total GC network as in our experiments. Moreover, situations where the differential GC is of more interest generally involve a small $K$ (\eg, brain signals collected under various symptom stages), and in cases of small $K$, this issue does not occur.

\section{Conclusion}

This paper proposed three sparse formulations named CGN, DGN and FGN for estimating multiple Granger causality networks with common causality structure across multiple time series and differential structures belonging to individual time series. These formulations can be applied to brain connectivity analysis where we are interested in a group-level inference and connectivity differences among subject conditions. The proposed formulations employed the group and fused lasso penalties with a weight prior to enhance the accuracy of estimating the sparsity of GC networks. The non-convex $\ell_{2,1/2}$ penalty was used to further improve the estimation in low-sample settings. The estimation problems were used in combination with the extended BIC as a model selection criterion which selected an optimal pair of penalty parameters, thus completing our scheme of learning multiple GC networks at optimal sparsity.

On average, our approaches improved F1 and FPR by 3-26\% and 0.6-13\%, respectively, over existing \emph{sparse multiple Granger graphical model} methods in literature. A main factor that determined CGN's accuracy was the density of the common ground-truth network, while DGN/FGN were slightly affected by the density of the differential ground-truth network. Contrary to previous results, DGN/FGN's accuracy was favorably insensitive to the number of models ($K$), and their performance improved relative to earlier methods even when $K$ was small. However, note that the number of variables grows linearly as $K$ increases, thus affecting the computational complexity in the algorithm's point of view.

We also used our formulations to analyze the differences of effective brain connectivity (in GC sense) between ADHD and TDC subjects with resting-state fMRI time series data obtained from the ADHD-200 dataset. Our formulations found results that were consistent with previous studies supported by both clinical and functional evidence from ADHD literature, asserting that the orbitofrontal and limbic system regions of the brain appeared highly related to ADHD.

\section{Acknowledgments}
This work was financially supported by the 90th Anniversary of Chulalongkorn University Fund (Ratchadaphiseksomphot Endowment Fund) and the 2020 Chula Engineering research grant. The first author gratefully acknowledges the support of the CUEE Master Honours Program Scholarship from the Department of Electrical Engineering, Chulalongkorn University.

\appendix
\section{Average performance metrics}
\label{sec:avg_performance}
The average performances of GC network estimation and comparisons to existing methods in literature are presented in \Cref{tab:classification_metric} with the classification metrics including the F1 score, false-positive rate (FPR), true positive rate (TPR), accuracy (ACC), and Matthews correlation coefficient (MCC). These results are obtained by following the experiment settings presented in Section 3.2-3.5 of the paper. Similar conclusions can be drawn when the averaged indicators are considered, in addition to using medians as shown in the paper. For all settings, both of our convex and non-convex formulations generally outperformed the comparative works when considering F1 score, FPR, and MCC, while all the methods had competing performances in terms of TPR and ACC. As the ground-truth GC networks were set in a sparse setting, the data were imbalanced with the majority of negatives (null GC). Therefore, when evaluated on data of different sizes, the improvement gains were most pronounced when using the F1 score, FPR (as the two indices do not consider TN) and MCC that is known for considering both of positive and negative predictions better than ACC. 

It is worth noting how our performances varied upon the density of ground-truth networks and the number of models. We note that our reported performances were contributed from both mathematical properties of the formulations and penalty selection using eBIC. For CGN, as the density increases, the F1 score and MCC were improved, likely due to a good selection of penalty by eBIC that resulted in estimated networks that contain less FNs relative to more FPs. The performances of DGN/FGN did not vary much when the differential density increased, except cvx-FGN. For DGN, unlike Skrip19b, our performances were not sensitive to the number of models and improved over other methods for both $K=5$ and $K=50$. Our arguments supporting these results follow the same discussion in the paper. 

\begin{table}
\caption{The performance average and standard deviation (in parentheses) over 100 runs of joint GC estimations. The bold-face values indicate the best performance among the comparison.}
\begin{subtable}[t]{1\textwidth}
\centering
\caption{\textbf{CGN} performance as the density of ground-truth common GC networks varied.}
\begin{tabular}{c|cccc|cccc|} \cline{2-9} 
&\multicolumn{4}{c|}{Common density: 10\%} & \multicolumn{4}{c|}{Common density: 20\%} \\ 
\cline{2-9}
 & CGN & cvx-CGN & Song17C & Greg15 & CGN & cvx-CGN & Song17C & Greg15 \\ 
\cline{2-9}
F1& \textbf{59.2} (4.4) & 57.7 (4.6) & 52.4 (5.8) & 52.6 (6.3) & 70.0 (2.3) & \textbf{70.9} (3.0) & 61.0 (4.9) & 59.6 (4.4) \\ 
FPR& \textbf{14.7} (2.7) & 15.7 (2.8) & 19.6 (4.4) & 19.4 (4.7) & 19.0 (2.0) & \textbf{18.1} (2.4) & 28.6 (6.0) & 30.3 (5.8) \\ 
TPR& \textbf{100.0} (0.0) & \textbf{100.0} (0.0) & 99.8 (0.8) & 99.7 (1.0) & \textbf{100.0} (0.0) & \textbf{100.0} (0.1) & 99.9 (0.5) & 99.8 (0.7) \\ 
ACC& \textbf{86.7} (2.4) & 85.8 (2.5) & 82.3 (4.0) & 82.4 (4.2) & 84.4 (1.6) & \textbf{85.1} (2.0) & 76.5 (4.9) & 75.2 (4.7) \\ 
MCC& \textbf{60.0} (4.1) & 58.6 (4.2) & 53.5 (5.4) & 53.7 (5.8) & 66.1 (2.4) & \textbf{67.1} (3.1) & 56.0 (5.5) & 54.4 (5.1) \\ 
\cline{2-9}
 \end{tabular} 
\end{subtable} \\

\begin{subtable}[t]{1\textwidth}
\centering
\caption{\textbf{DGN} performance as the density of ground-truth common GC networks varied.}
\begin{tabular}{c|cccc|cccc|} \cline{2-9} 
&\multicolumn{4}{c|}{Differential density: 1\%} & \multicolumn{4}{c|}{Differential density: 5\%} \\ 
\cline{2-9}
 & DGN & cvx-DGN & Song17D & Skrip19b & DGN & cvx-DGN & Song17D & Skrip19b \\ 
\cline{2-9}
F1& 95.1 (2.0) & \textbf{95.6} (1.9) & 90.6 (3.1) & 82.4 (2.2) & 95.3 (1.8) & \textbf{95.6} (1.7) & 84.1 (2.3) & 68.9 (2.0) \\ 
FPR& 1.0 (0.5) & \textbf{0.8} (0.5) & 1.6 (0.7) & 4.9 (0.8) & 1.4 (0.7) & \textbf{1.2} (0.6) & 4.7 (1.3) & 14.3 (1.4) \\ 
TPR& 98.0 (1.5) & 97.5 (1.7) & 94.0 (3.7) & \textbf{99.6} (0.4) & \textbf{99.0} (0.7) & 98.4 (1.0) & 93.7 (3.4) & 98.8 (0.8) \\ 
ACC& 98.9 (0.5) & \textbf{99.1} (0.4) & 98.0 (0.7) & 95.5 (0.7) & 98.6 (0.5) & \textbf{98.7} (0.5) & 95.0 (0.9) & 87.5 (1.1) \\ 
MCC& 94.6 (2.2) & \textbf{95.2} (2.1) & 89.6 (3.4) & 81.6 (2.2) & 94.6 (2.0) & \textbf{94.9} (1.9) & 81.9 (2.5) & 66.7 (2.0) \\ 
\cline{2-9}
 \end{tabular} 
\end{subtable} \\
 
\begin{subtable}[t]{1\textwidth}
\centering
\caption{\textbf{DGN} performance as the number of models varied.}
\begin{tabular}{c|cc|cc|cc|cc|} \cline{2-9}  
&\multicolumn{2}{c|}{DGN} & \multicolumn{2}{c|}{cvx-DGN} & \multicolumn{2}{c|}{Song17D} & \multicolumn{2}{c|}{Skrip19b} \\  \cline{2-9}
 & $K=5$ & $K=50$ & $K=5$ & $K=50$ & $K=5$ & $K=50$ & $K=5$ & $K=50$ \\ 
\cline{2-9}
F1& 95.3 (1.8) & 96.1 (1.2) & 95.6 (1.7) & 95.3 (0.9) & 84.1 (2.3) & 82.2 (1.9) & 68.9 (2.0) & 82.9 (1.6) \\ 
FPR& 1.4 (0.7) & 1.0 (0.5) & 1.2 (0.6) & 1.2 (0.5) & 4.7 (1.3) & 5.3 (1.6) & 14.3 (1.4) & 5.5 (0.8) \\ 
TPR& 99.0 (0.7) & 98.4 (0.8) & 98.4 (1.0) & 97.6 (1.0) & 93.7 (3.4) & 92.3 (4.0) & 98.8 (0.8) & 94.7 (0.9) \\ 
ACC& 98.6 (0.5) & 98.9 (0.4) & 98.7 (0.5) & 98.7 (0.3) & 95.0 (0.9) & 94.4 (0.9) & 87.5 (1.1) & 94.5 (0.7) \\ 
MCC& 94.6 (2.0) & 95.5 (1.3) & 94.9 (1.9) & 94.6 (1.1) & 81.9 (2.5) & 79.7 (1.9) & 66.7 (2.0) & 80.6 (1.8) \\ 
\cline{2-9}
 \end{tabular} 
\end{subtable} \\

\begin{subtable}[t]{1\textwidth}
\centering
\caption{\textbf{FGN} performance as the density of ground-truth common GC networks varied.}
\begin{tabular}{c|cccc|cccc|} \cline{2-9} 
&\multicolumn{4}{c|}{Differential density: 1\%} & \multicolumn{4}{c|}{Differential density: 5\%} \\ 
\cline{2-9}
 & FGN & cvx-FGN & Song15 & Skrip19a & FGN & cvx-FGN & Song15 & Skrip19a \\ 
\cline{2-9}
F1& \textbf{95.8} (3.0) & 89.2 (5.8) & 85.0 (3.5) & 92.5 (3.0) & \textbf{95.8} (2.9) & 94.3 (3.1) & 83.4 (3.4) & 92.3 (1.5) \\ 
FPR& \textbf{1.0} (0.8) & 2.9 (1.7) & 2.8 (1.1) & 1.4 (0.6) & \textbf{1.3} (1.1) & 1.8 (1.2) & 5.1 (1.7) & 2.1 (0.5) \\ 
TPR& \textbf{99.5} (0.5) & \textbf{99.5} (0.5) & 91.7 (3.7) & 97.1 (2.1) & \textbf{99.4} (0.5) & 98.9 (0.8) & 93.8 (2.4) & 97.4 (1.4) \\ 
ACC& \textbf{99.1} (0.7) & 97.4 (1.5) & 96.6 (1.0) & 98.4 (0.6) & \textbf{98.8} (0.9) & 98.3 (1.0) & 94.7 (1.4) & 97.8 (0.4) \\ 
MCC& \textbf{95.5} (3.2) & 88.5 (5.9) & 83.5 (3.8) & 91.8 (3.3) & \textbf{95.2} (3.2) & 93.5 (3.4) & 81.1 (3.8) & 91.3 (1.6) \\ 
\cline{2-9}
 \end{tabular} 
\end{subtable} \\
 
\begin{subtable}[t]{1\textwidth}
\centering
\caption{Comparison of non-convex versus convex formulations in a low-sample setting.}
\begin{tabular}{c|cc|cc|cc|} \cline{2-7} 
 & CGN & cvx-CGN & DGN & cvx-DGN & FGN & cvx-FGN \\ 
\cline{2-7}
F1& \textbf{76.6} (5.0) & 54.1 (5.9) & \textbf{88.5} (4.1) & 72.9 (4.7) & \textbf{88.5} (5.4) & 73.3 (6.7) \\ 
FPR& \textbf{5.2} (2.0) & 18.1 (4.8) & \textbf{0.4} (0.2) & 3.3 (1.1) & \textbf{0.4} (0.3) & 3.5 (1.9) \\ 
TPR& 92.0 (6.4) & \textbf{98.7} (2.1) & \textbf{81.4} (6.7) & 69.2 (7.8) & \textbf{81.9} (9.0) & 70.5 (10.3) \\ 
ACC& \textbf{94.5} (1.6) & 83.5 (4.2) & \textbf{97.1} (0.9) & 92.8 (1.1) & \textbf{97.1} (1.2) & 92.9 (1.7) \\ 
MCC& \textbf{75.2} (5.0) & 54.8 (5.6) & \textbf{87.4} (4.1) & 69.2 (5.0) & \textbf{87.5} (5.4) & 69.8 (7.4) \\ 
\cline{2-7}
\end{tabular} 
\end{subtable}
\label{tab:classification_metric}
\end{table}

\section{Overlapped penalization of DGN}
\label{sec:overlapped_dgn}

\begin{figure}[ht]
\begin{subfigure}[b]{0.45\linewidth}
\includegraphics[width=\linewidth]{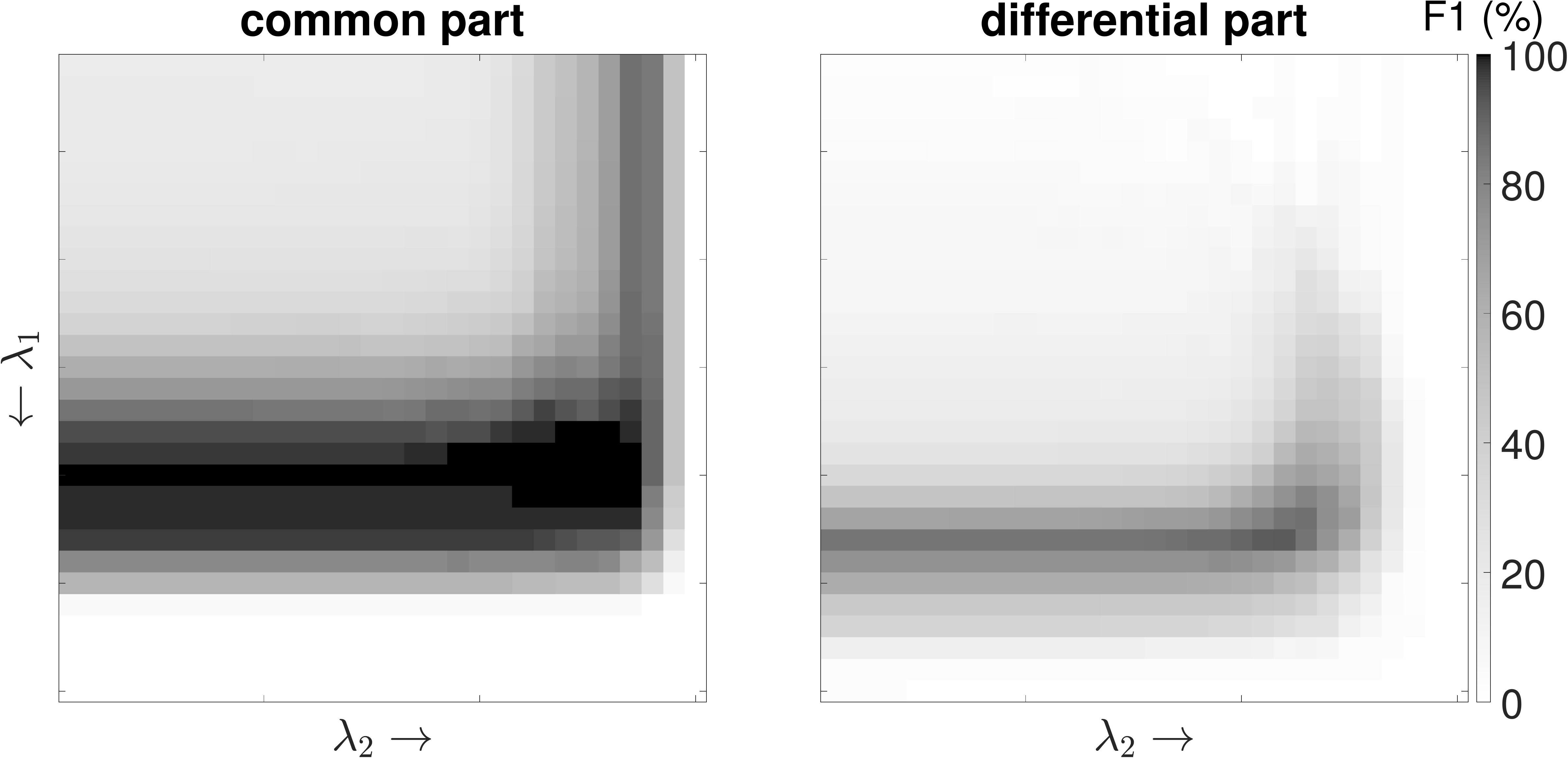}
\caption{$K=5$.}
\label{fig:K5_supp}
\end{subfigure}
\begin{subfigure}[b]{0.45\linewidth}
\includegraphics[width=\linewidth]{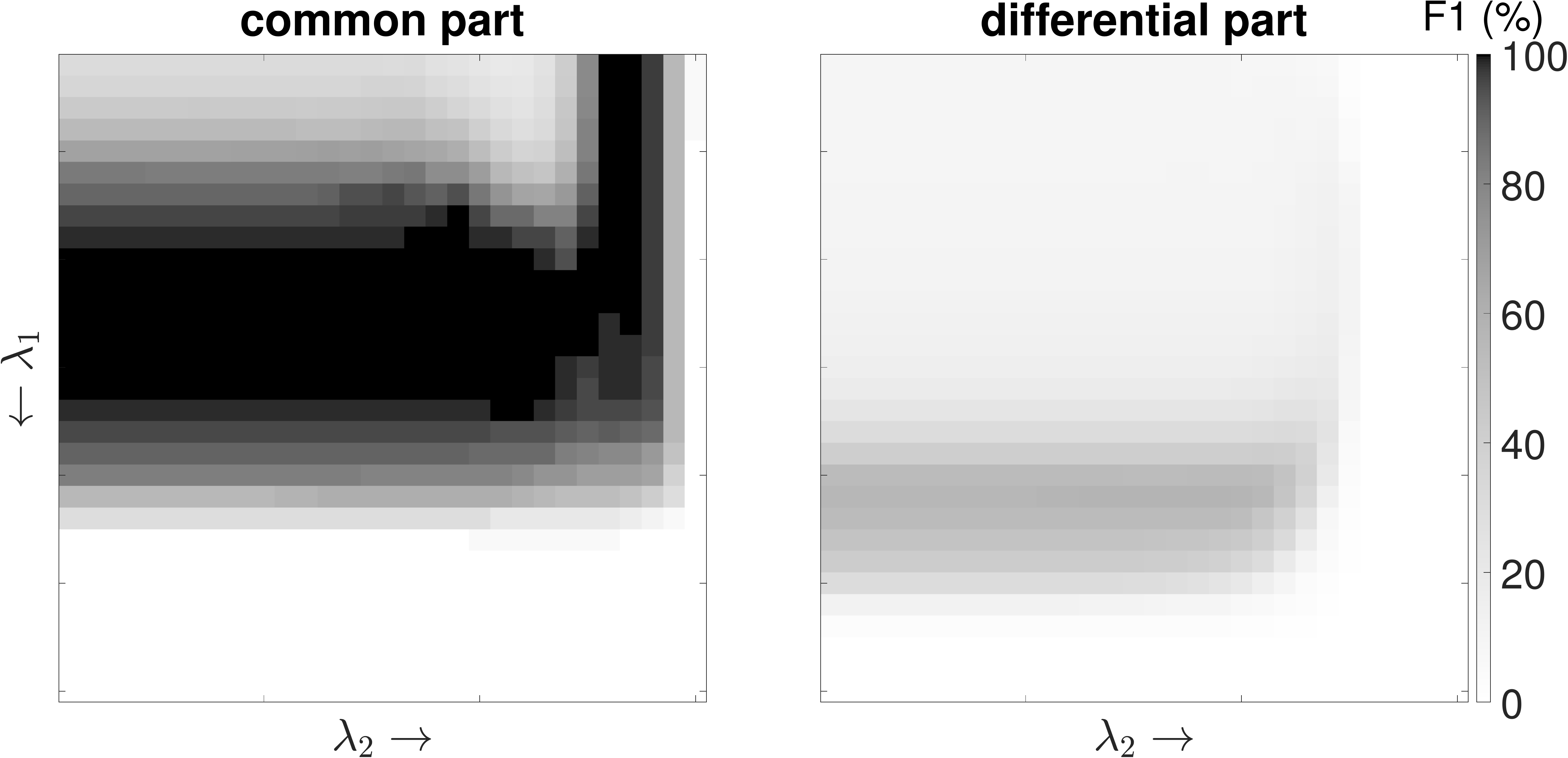}
\caption{$K=50$.}
\label{fig:K50_supp}
\end{subfigure}
\caption{The F1 scores as $(\lambda_1,\lambda_2)$ varied when evaluated on the common and differential parts of the estimated GC networks. The F1 scores were averaged over 100 data realizations where the darker color indicates the higher F1 score.}
\label{fig:dgn_overlapped_penalty}
\end{figure}

The DGN formulation requires a selection of the two penalty parameters, $(\lambda_1,\lambda_2)$ in the two nested penalties to control the sparsity of the common and differential networks. As discussed in the limitations of our method that when evaluating the performance metrics on the common and differential networks separately, the best performances occurred at different pairs of $(\lambda_1,\lambda_2)$. This section shows an additional result in \Cref{fig:dgn_overlapped_penalty} to support this argument. When $K=50$, high F1 scores evaluated on the common part were obtained in different regions of $(\lambda_1,\lambda_2)$ from those evaluated on the differential part. This difference appeared in less degree for $K=5$. We note that all reported performances in the paper were evaluated on the total network. Hence, achieving the best performance of both common and differential networks is not quite possible when the number of models is relatively large. 

\section{Vector formulations of joint GC estimation}
\label{sec:vector_formulation}
The CGN, DGN and FGN formulations can be presented in a unified framework as 
\begin{equation}
\underset{A^{(1)},\ldots,A^{(K)}}{\minimize} ~ \frac{1}{2N}\sum_{k=1}^{K} \left \Vert Y^{(k)}-A^{(k)}H^{(k)} \right \Vert _{F}^{2}+g(A^{(1)},\ldots, A^{(K)}),
\label{eq:jointGC2}
\end{equation}
with variables $A^{(k)}$ for $k=1,\ldots,K$. The problem parameters are $N=T-p$, and 
\[
Y^{(k)} = \begin{bmatrix} y^{(k)}(p+1) &\cdots&y^{(k)}(T) \end{bmatrix}, \quad 
H^{(k)} = \begin{bmatrix}
y^{(k)}(p) & y^{(k)}(p+1) & \cdots & y^{(k)}(T-1) \\
\vdots & \vdots & \ddots & \vdots\\
y^{(k)}(2) & y^{(k)}(3) & \cdots & y^{(k)}(T-p+1) \\
y^{(k)}(1) & y^{(k)}(2) & \cdots & y^{(k)}(T-p) 
\end{bmatrix}.
\]
In order to describe the main algorithms for solving \eqref{eq:jointGC2}, and for the ease of mathematical presentation, we reformulate the problem in a vector format described as 
\begin{equation}
\minimize \quad \frac{1}{2N} \Vert Gx-b \Vert_{2}^{2} + g(x),
\label{eq:vec_formulation1}
\end{equation}
with variable $x \in \reals^{\tilde{n}}$ and problem parameters $G \in \reals^{\tilde{m} \times \tilde{n}}, b \in \reals^{\tilde{m}}$. The $\Vert \cdot \Vert_2^2$ term represents the least-squares objective (model fitting) and $g$ is the penalty function of the choices in CGN, DGN, and FGN. The optimization variable $x$ refers to the $n$-dimensional $p$-order VAR model parameters of $K$ models: $A^{(k)}$ for $k=1,\ldots, K$. Specifically, each model's VAR parameter contains $p$-lag coefficients as $A^{(k)} = \begin{bmatrix} A_1^{(k)} & A_2^{(k)} & \cdots & A^{(k)}_p \end{bmatrix}$, and each lag coefficient is an $n \times n$ matrix, \ie,  $A_{p}^{(k)} \in \reals^{n \times n}$. For each model, our regularization technique requires grouping all VAR-lag coefficients into
\begin{equation}
B^{(k)}_{ij}=\begin{bmatrix}
(A^{(k)}_{1})_{ij} & \cdots & (A^{(k)}_{p})_{ij}
\end{bmatrix} \in \reals^p.
\label{eq:bij}
\end{equation}
In addition, the regularization involves grouping the parameters across models in order to enforce a common network among them, so we pool $B_{ij}^{(k)}$ from $K$ models to
\begin{equation}
C_{ij} = \begin{bmatrix}
B_{ij}^{(1)} & B_{ij}^{(2)} & \cdots & B_{ij}^{(K)} 
\end{bmatrix} \in \reals^{pK}.
\label{eq:Cij}
\end{equation}
The optimization variable, $x$, is the vector obtained by pooling $C_{ij}$ from all $ 1 \leq i,j \leq n$, that is, 
\begin{equation}
x = (C_{11}, \ldots,C_{1n},C_{21},\ldots,C_{2n}, \ldots,C_{n1},\ldots, C_{nn}).
\label{eq:vecx}
\end{equation}
It is clear that the dimension of $x$ is then $\tilde{n} = pKn^2$, and that $x$ can be partitioned into two different ways. When splitting $x$ to blocks of size $pK$, each block is $C_{ij}$, and when the block size is $p$, each block of $x$ is $B_{ij}^{(k)}$. 

\begin{figure}[ht]
\centering
\begin{subfigure}[b]{0.22\linewidth}
    \centering
    \includegraphics[width=1\linewidth]{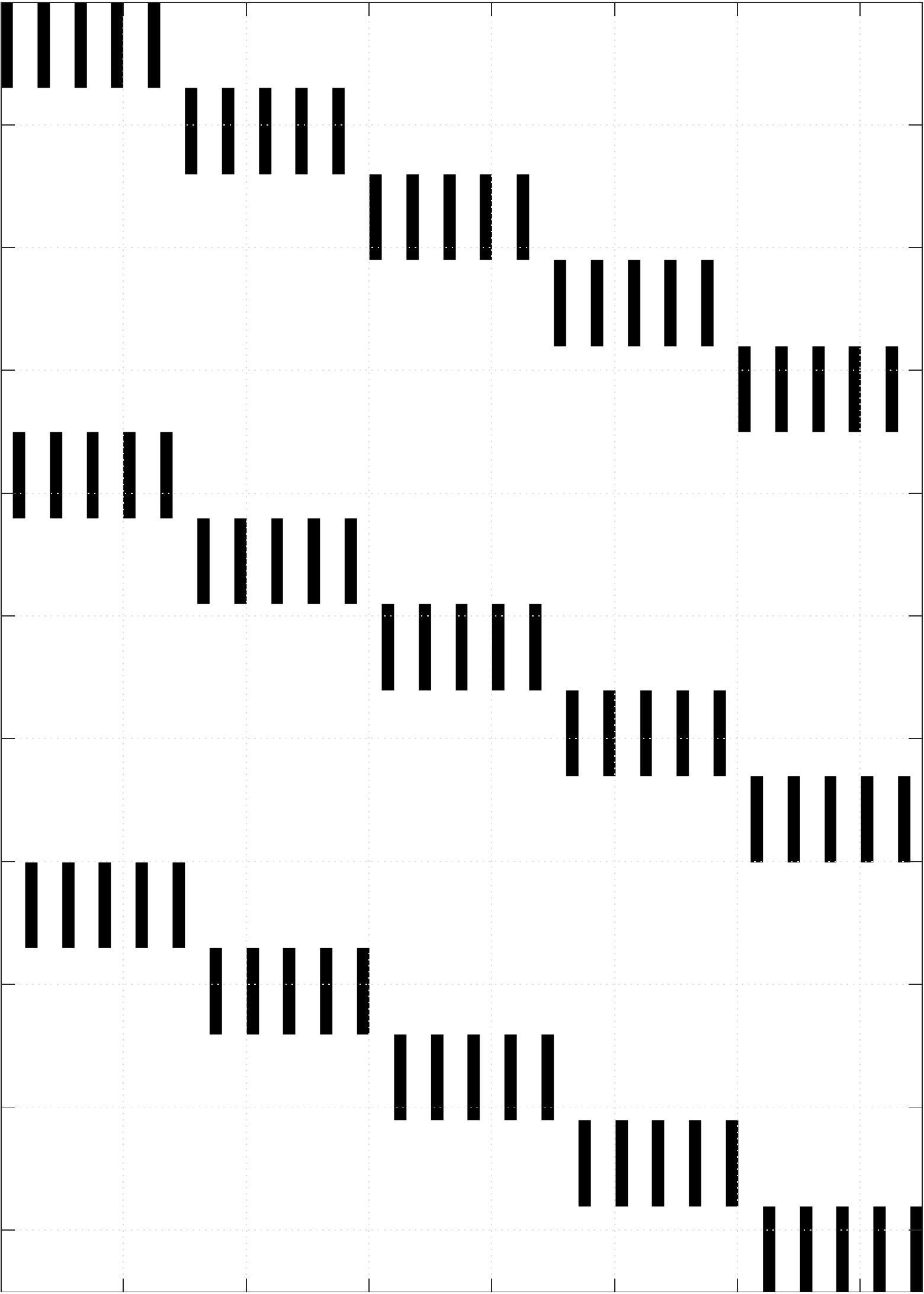}
    \caption{$G$}
    \label{fig:G_sparsity}
\end{subfigure}
\begin{subfigure}[b]{0.31\linewidth}
    \centering
    \includegraphics[width=1\linewidth]{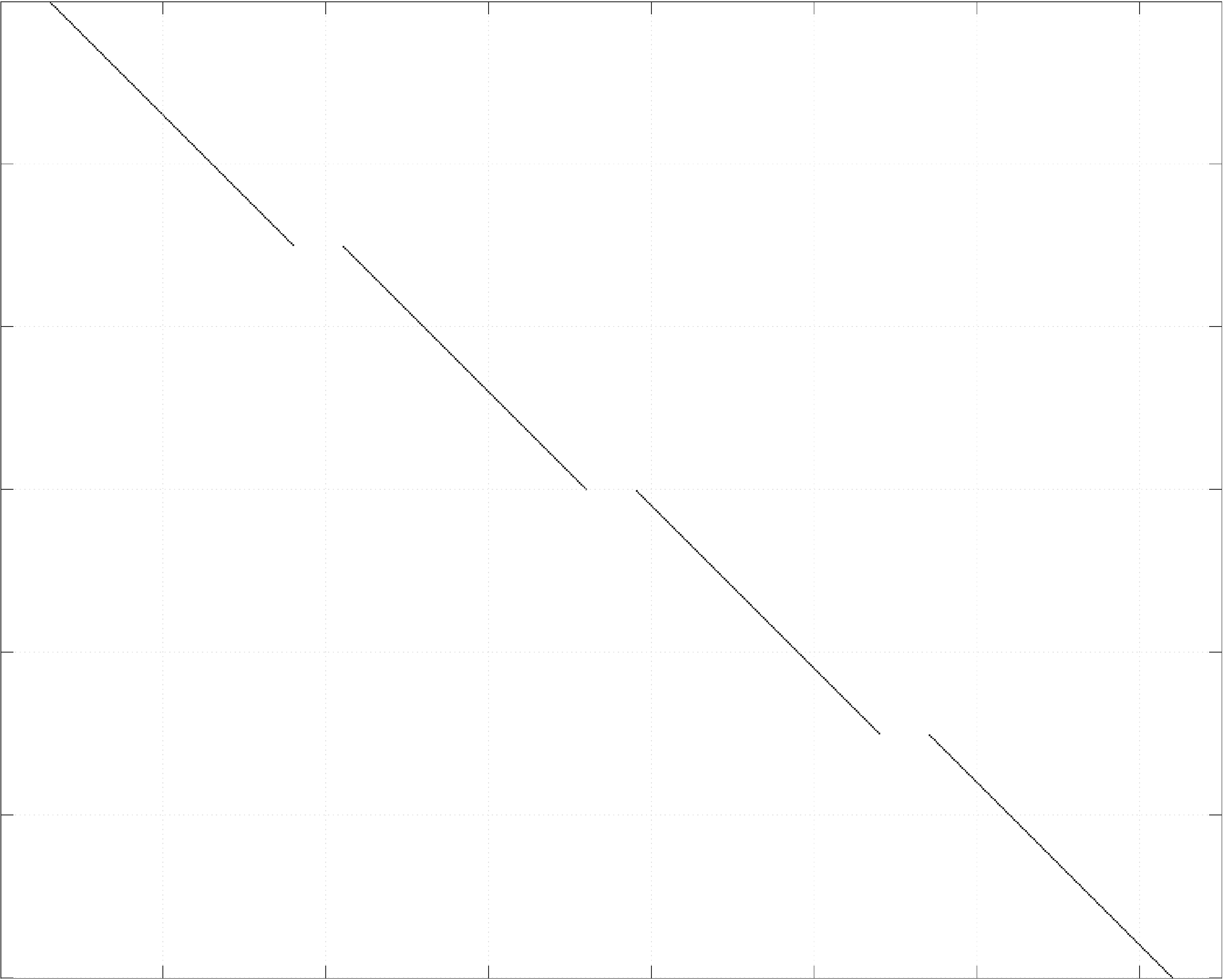}
    \caption{$P$}
    \label{fig:P_sparsity}
\end{subfigure}
\begin{subfigure}[b]{0.22\linewidth}
    \centering
    \includegraphics[width=1\linewidth]{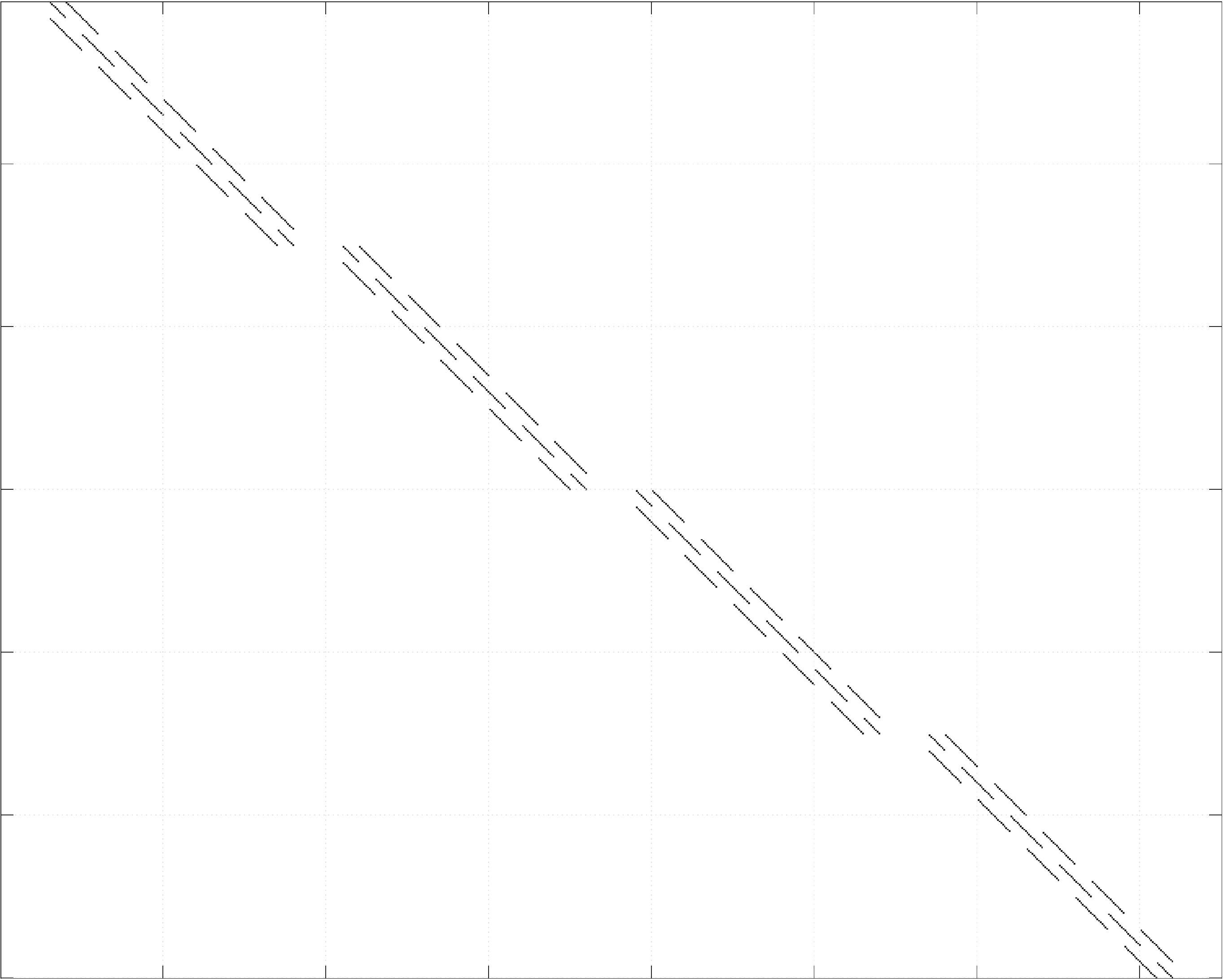}
    \caption{$D$}
    \label{fig:D_sparsity}
\end{subfigure} \\
\begin{subfigure}[b]{0.22\linewidth}
    \centering
    \includegraphics[width=1\linewidth]{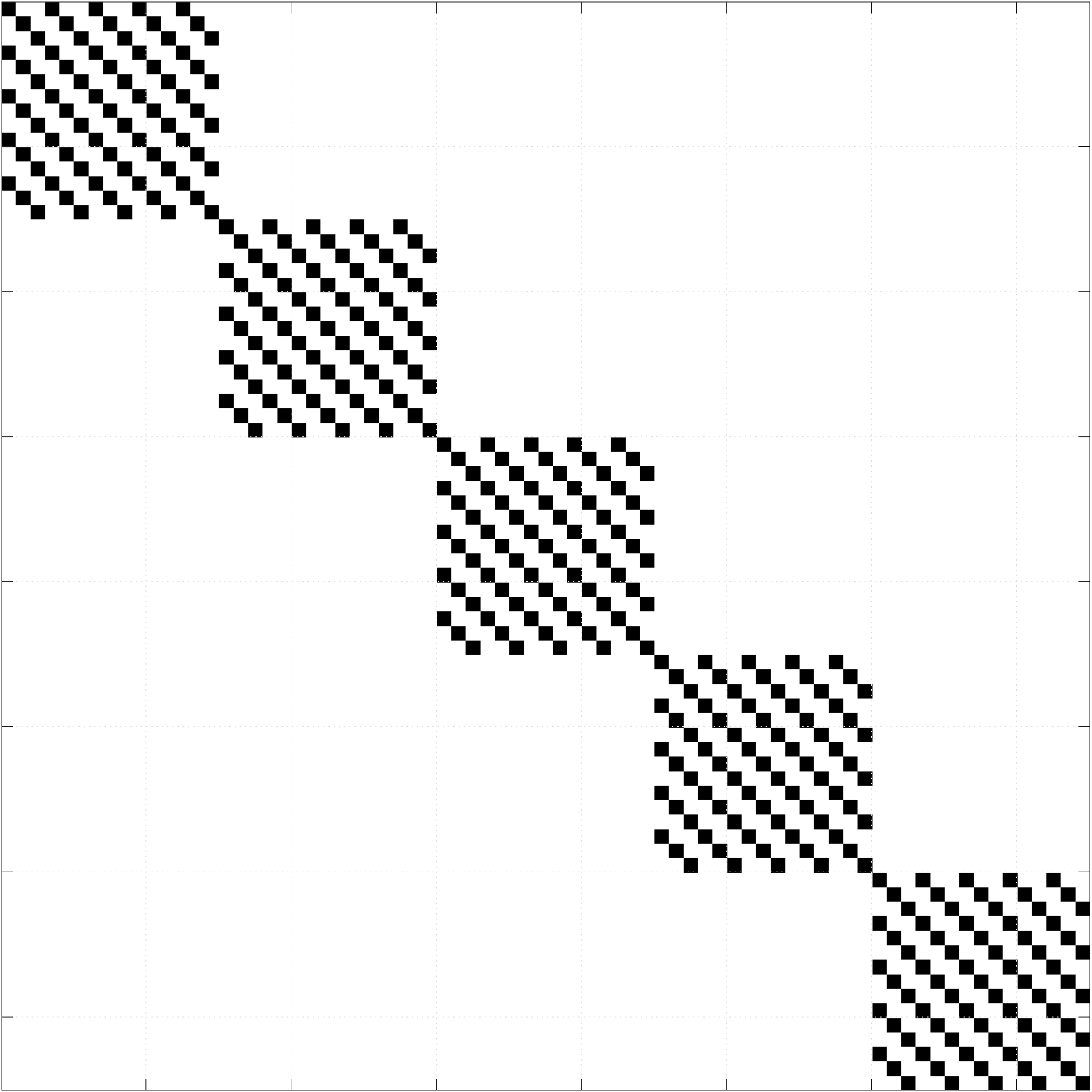}
    \caption{$G^{T}G$}
    \label{fig:GtG_sparsity}
\end{subfigure}
\begin{subfigure}[b]{0.22\linewidth}
    \centering
    \includegraphics[width=1\linewidth]{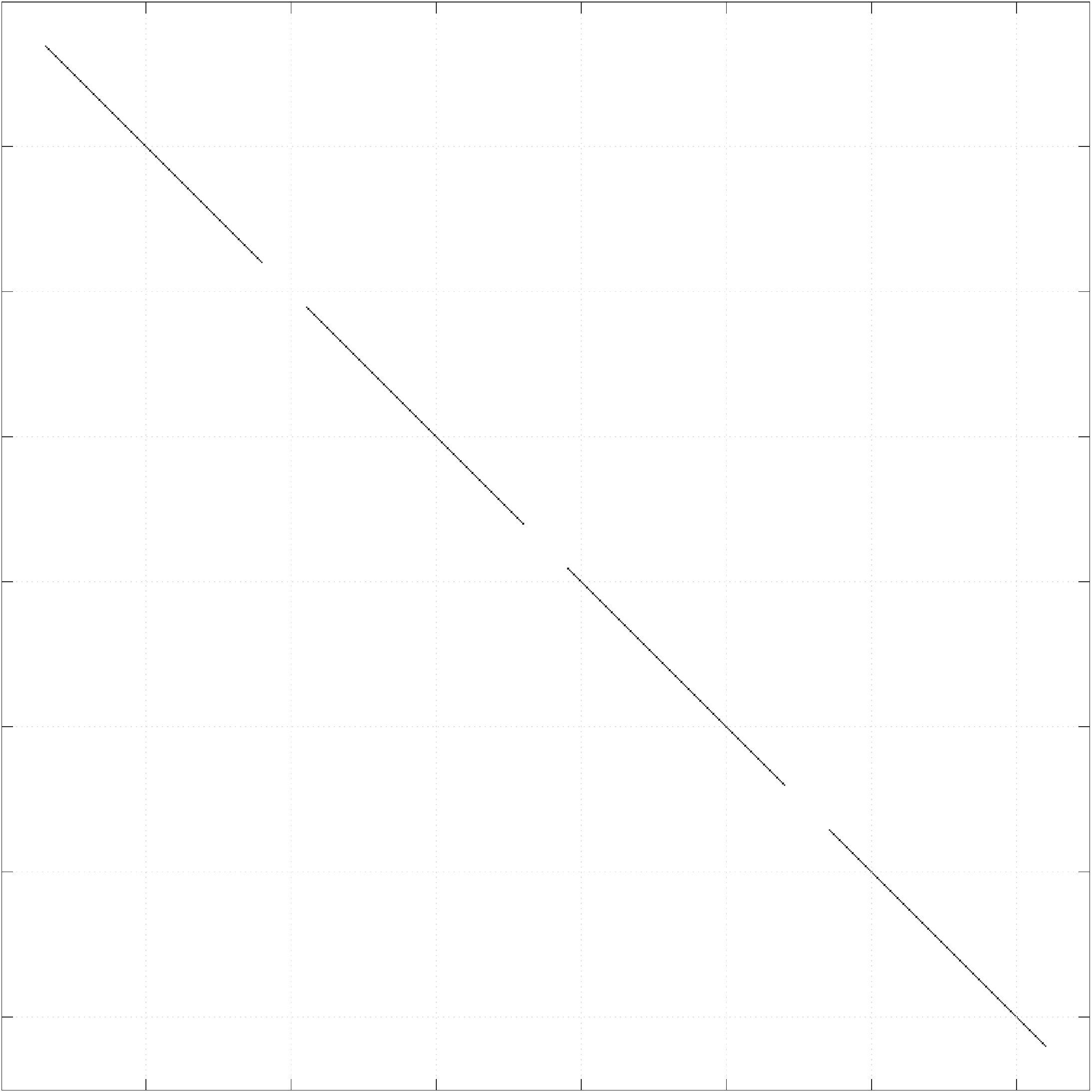}
    \caption{$P^{T}P$}
    \label{fig:PtP_sparsity}
\end{subfigure}
\begin{subfigure}[b]{0.22\linewidth}
    \centering
    \includegraphics[width=1\linewidth]{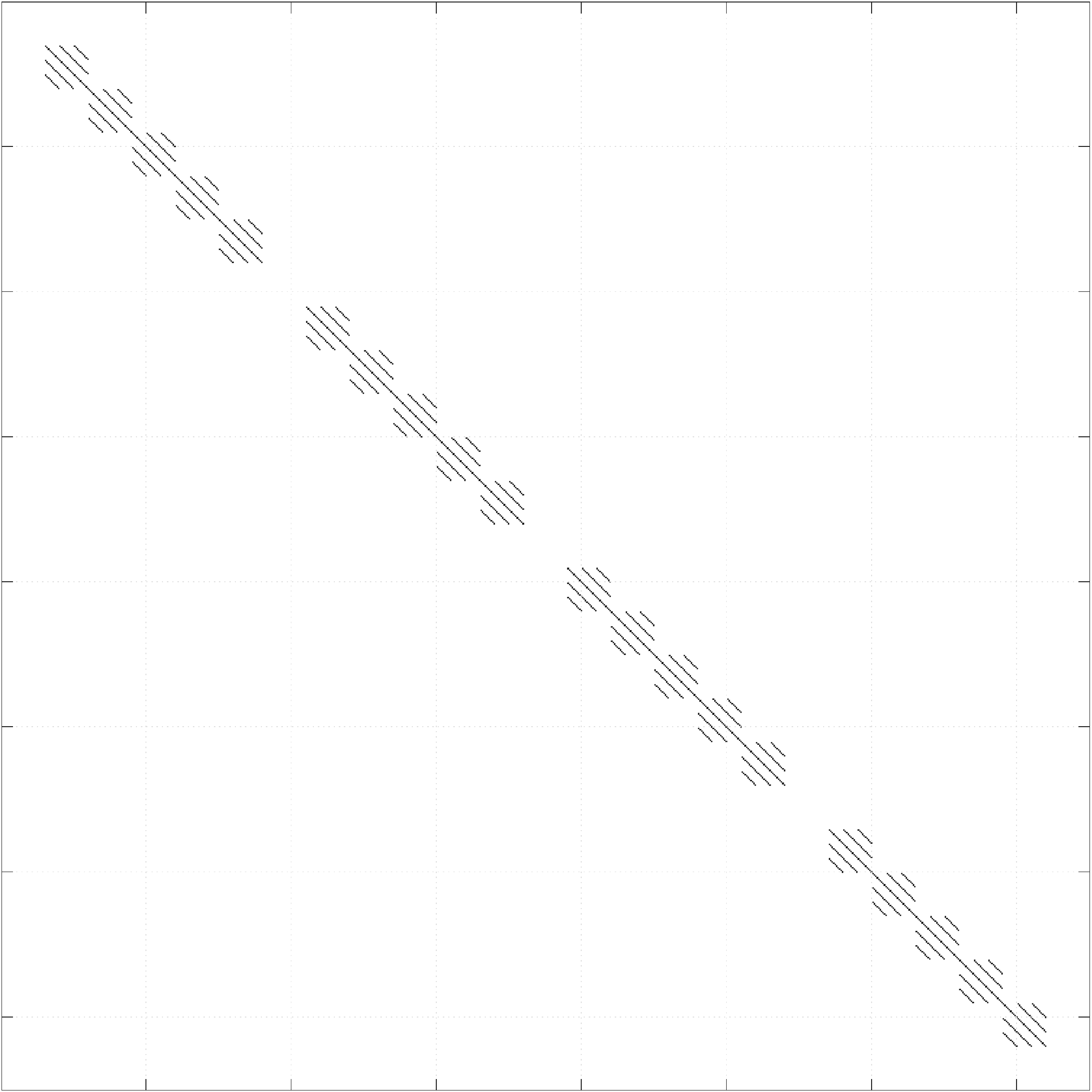}
    \caption{$D^{T}D$}
    \label{fig:DtD_sparsity}
\end{subfigure}
\caption{Sparsity pattern of parameter-matrices involved in the algorithm: $n=5, p=10, K=3$. The size of involved matrices are $G$: $nNK \times n^2pK$, $P$: $(n^2-n)pK \times n^2pK$, and $D$: $ {K \choose 2}  (n^2-n)p \times n^2pK$. }
\label{fig:matrix_sparsity}
\end{figure}

From the definition of $x$ in \eqref{eq:vecx}, the problem parameters, $b$ and $G$, in \eqref{eq:vec_formulation1} are obtained by matching $\Vert Gx-b \Vert_{2}^{2} := \sum_{k=1}^{K} \Vert Y^{(k)}-A^{(k)}H^{(k)} \Vert_{F}^{2}$, which results in $G$ of size $\tilde{m} \times \tilde{n}$ with $\tilde{m} = nNK$ and some sparse structure shown in \Cref{fig:G_sparsity}. The matrix $G$ is concatenated by $K$ row-blocks; each of which is a block diagonal matrix. As shown in \Cref{fig:GtG_sparsity}, the structure of $G$ gives a block-diagonal form of $G^TG$, which is required in our algorithm.

The regularizations in the three formulations only penalize the off-diagonal terms of VAR coefficients (as we do not analyze a Granger cause from one variable to itself), the penalty function $g$ requires a projection mapping $x$ to the subspace of off-diagonal entries VAR parameters first. Moreover, the regularization involves grouping the model parameters over either all of $p$-lag coefficients or all of $K$ models, hence, it is convenient to define a weighted group-norm penalty associated with a partition $\mathcal{B}$ of $x$ as
\begin{equation}
h(x;\mathcal{B}) = \sum_{k \in \mathcal{B}} w_{k}\Vert x_{k} \Vert _{2}^{q}
\label{eq:weighted_group_norm}
\end{equation}
where $\mathcal{B}$ is a partition of $\{1,2,\ldots,\tilde{n} \}$ \footnote{The collection $\mathcal{B} = \{c_1,c_2,\ldots,c_L \}$ is a partition of $\{1,2,\ldots,\tilde{n} \}$ if $\bigcup \mathcal{B} = \{1,2,\ldots, \tilde{n} \}$ and $c_i \bigcap c_j = \varnothing$ for $i \neq j$.}, $w_k > 0$, and $x_{k}$ is the $k^\mathrm{th}$ partition of $x$. We note that $h$ is, in fact, parametrized by the choices of weight $w_k$ and partition $\mathcal{B}$, but to keep the notation simple, we express $h(x;\mathcal{B})$ without $w_k$. It is also important to note that the partion $\mathcal{B}$ has to implicitly agree with the argument of $h(\cdot \; ; \mathcal{B})$; for example, the partition $\mathcal{B}$ used in $h(Lx;\mathcal{B})$ must be associated with the index set $\{1,2,\ldots,l \}$ where $l$ is the dimension of $Lx$.

Two partitions are involved in our problem. Suppose $h$ is evaluated with an input argument that is a vector $u$ of size $l$. We define $\mathcal{P}$ and $\mathcal{K}$, the two partitions that correspond to splitting $u$ into blocks of size $p$ and $pK$, respectively.
\begin{gather*}
\mathcal{P} = \left \{ \{1,2,\ldots,p \}, \{p+1,p+2,\ldots,2p\}, \ldots,\{ l-p+1,l-p+2,\ldots, l \} \right \}, \\
\mathcal{K} = \left \{ \{1,2,\ldots,pK \}, \{pK+1,pK+2,\ldots,2pK\}, \ldots,\{ l-pK+1,l-pK+2,\ldots, l \} \right \}.
\end{gather*}
These allow us to state the penalty function $g$ of all formulations as follows.
\begin{eqnarray}
\text{CGN} &:& g(x) =  \lambda_1 h( Px; \mathcal{K}) \label{eq:penalty_cgn} \\
\text{DGN} &:& g(x) =  \lambda_1 h(P x; \mathcal{P}) + \lambda_2 h( P x ;  \mathcal{K} )  \label{eq:penalty_dgn} \\
\text{FGN} &:& g(x) = \lambda_1 h( P x; \mathcal{P} ) + \lambda_2 h( D x; \mathcal{P}) \label{eq:penalty_fgn} 
\end{eqnarray}

It is worth noting that each $h$ may have different pre-defined weight $w_{l}$ for each block and the length of input vector may be not the same (\eg~$Dx$ and $Px$ are of different sizes in FGN, thus, $\mathcal{P}$ in the two terms of FGN are not exactly the same but they both correspond to partitions of size $p$). The matrix $P$ corresponds to the projection of all entries of VAR coefficients to the off-diagonal entries, hence $P$ has the size of $(n^2-n)pK \times n^{2}pK$. To illustrate this, consider the case of $n=2$ and then $x = (C_{11},C_{12},C_{21}, C_{22})$ with $C_{ij} \in \reals^{pK}$. To extract only off-diagonal entries of all-lag VAR coefficients, or equivalently, $Px = (C_{12},C_{21})$, the projection matrix $P$ is given by
\[
P = \begin{bmatrix} 0 & I_{pK} & 0 & 0 \\ 0 & 0 & I_{pK} & 0 \end{bmatrix}.    
\]
The projection $P$ is fat and sparse, and always contains some entire zero columns; see its sparsity example in \Cref{fig:P_sparsity}.

As for the FGN formulation, the term $Dx$ is for the fused lasso that penalizes the differences between parameters of any two models. Thus, $D$ maps VAR coefficients of all models into the difference of off-diagonal VAR coefficients between any two models. This can be considered as a composite of the difference transformation with the projection transformation. As an example of $K=3$ and for a fixed $(i,j)$, the term $Dx$ is desired to represent 
\begin{equation}
\begin{bmatrix}
B_{ij}^{(1)}-B_{ij}^{(2)}\\B_{ij}^{(1)}-B_{ij}^{(3)}\\B_{ij}^{(2)}-B_{ij}^{(3)}
\end{bmatrix}
=
\begin{bmatrix}
I_{p} & -I_{p} & 0 \\ I_{p} & 0 & -I_{p} \\ 0 & I_{p} & -I_{p}
\end{bmatrix}
\begin{bmatrix} B_{ij}^{(1)}\\B_{ij}^{(2)}\\B_{ij}^{(3)} \end{bmatrix} 
\triangleq \left ( \underbrace{\begin{bmatrix} 1 & -1 & 0 \\ 1 & 0 & -1 \\ 0 & 1 & -1 \end{bmatrix}}_{D_K} \otimes I_p \right ) 
\begin{bmatrix} B_{ij}^{(1)}\\B_{ij}^{(2)}\\B_{ij}^{(3)} \end{bmatrix}.
\label{eq:Dx1}
\end{equation}
Therefore, $Dx$ is a concatenation of the vector in \eqref{eq:Dx1} when $(i,j)$ is varied for all $ 1 \leq i,j \leq n$ and $i \neq j$. For a general $K$, we see that $D_K$ takes all possible differences between any two entries, so the structure of $D_K$ depends on $K$ and its dimension is $ {K \choose 2} \times K$. Define $\tilde{D}$ as the $(n^2-n)$-block diagonal matrix with all blocks of 
$D_K \otimes I_p$. Suppose $z = Px $ then $z$ contains only off-diagonal entries of VAR parameters. If $z$ is partitioned to blocks of size $p$, \ie, each block is $B^{(k)}_{ij}$ as in \eqref{eq:Dx1}, then we see that $ \tilde{D} z$ is $Dx$ as desired. Mathematically, $D = \tilde{D} P$ and has the size of $  {K \choose 2}  (n^2-n)p \times n^2pK$, but this expression should not be used in the numerical construction of $D x$. It can be tall, fat or square, depending on $(n,K)$; see a sparsity structure in \Cref{fig:D_sparsity}.

\section{Algorithms}
\label{sec:alg}
The penalty $g$ of CGN as in \eqref{eq:penalty_cgn} has one penalty term, while FGN and DGN have two group-norm penalties \eqref{eq:penalty_dgn}-\eqref{eq:penalty_fgn}; the latter two formulations can be more involved when finding a suitable algorithm for the non-convex setting. For non-convex non-smooth problems, existing proximal-type algorithms with a convergence often require the Kurdyka - \L{}ojasiewicz (KL) property of the cost function, which can be shown to be satisfied for quadratic and $\ell_{2,1/2}$-norm functions (as in our case) in \cite{FYW20}. To all formulations, the non-monotone accelerated proximal gradient algorithm (nmAPG)~\cite{LL15} and the proximal gradient method \cite{ABS13} can be directly applied with a convergence where the convergence rate of nmAPG was reported to be superior than the proximal gradient method. A common ground of these two algorithms is that it involves computing the proximal operator of $g$. As pointed out that $g$ of CGN has only one term and that $g$ is a group-norm; therefore, its proximal operator can be computed efficiently. On the other hand, when $g$ has two penalty terms and is a composite of group-norm with affine transformations as in \eqref{eq:penalty_dgn}-\eqref{eq:penalty_fgn}, the proximal step does not have almost-closed-form expression and can be computationally expensive. Therefore, we do not recommend applying nmAPG or the proximal gradient method to (non-convex) FGN and DGN.

As an alternative that works well in our implementation, we will present the ADMM algorithm with adaptive algorithm parameter with a note that its convergence is assured for convex formulations. However, for non-convex problems, we suggest an adaptive rule of the algorithm parameter that often gives convergent primal and dual residuals in practice. 

We present algorithms for solving the vector formulation presented in \Cref{sec:vector_formulation} without the scaling $N$.
\begin{equation}
\minimize \quad  (1/2) \Vert Gx-b \Vert_{2}^{2} + \lambda_{1} h(L_{1}x ; \mathcal{B}_1) + \lambda_{2} h (L_{2}x; \mathcal{B}_2),
\label{eq:vec_formulation2}
\end{equation}
with variable $x \in \reals^{\tilde{n}}$ and the problem parameters are $G \in \reals^{\tilde{m} \times \tilde{n}}, b \in \reals^{\tilde{m}},L_1 \in \reals^{\tilde{n}_1 \times \tilde{n}}$, and $L_2 \in \reals^{\tilde{n}_1 \times \tilde{n}}$. The dimensions of $L_1$ and $L_2$ and the partitions $\mathcal{B}_1, \mathcal{B}_2$ are up to the choice of penalty functions in each formulation as in \eqref{eq:penalty_cgn}-\eqref{eq:penalty_fgn}. We present the ADMM algorithm to solve \eqref{eq:vec_formulation2}, so it can be arranged into the ADMM format as 
\begin{equation}
\begin{array}{ll}
\minimize  &  f(x) + \tilde{g}(z),\\
\mbox{subject to} &  Ax+Bz=0\\
\end{array}
\label{eq:ADMMformat}
\end{equation}
with variables $x \in \reals^{\tilde{n}}$ and $z = (z_1,z_2) \in \reals^{\tilde{n}_1} \times \reals^{\tilde{n}_2}$. We split $x$ and $z$ such that $z_1 = L_1x$ and $z_2 = L_2 x$. The function $f(x) = (1/2) \Vert Gx- b\Vert_2^2$ is convex and its gradient is Lipschitz continous. With the choice of splitting, we re-define the penalty $g$ from \eqref{eq:penalty_cgn}-\eqref{eq:penalty_fgn} as $\tilde{g}: \reals^{\tilde{n}_1} \times \reals^{\tilde{n}_1} \rightarrow \reals$, $\tilde{g}(z) = \lambda_{1} h(z_1 ; \mathcal{B}_1) + \lambda_{2} h (z_2; \mathcal{B}_2)$. Our ADMM format \eqref{eq:ADMMformat} corresponds to
\begin{equation}
B=-I, \; A = \begin{bmatrix} L_{1}\\ L_{2} \end{bmatrix},\; \text{where} \begin{bmatrix} L_{1}\\ L_{2} \end{bmatrix} = \begin{bmatrix} P \\ P \end{bmatrix}  \;\text{for DGN,  and}\;
\begin{bmatrix} L_{1}\\ L_{2} \end{bmatrix} = \begin{bmatrix} P \\ D \end{bmatrix} \;\text{for FGN.}
\label{eq:admm_AB}
\end{equation}
Following the algorithm description in \cite[\S 3]{BPC11}, the augmented Lagrangian is $ L_\rho(x,z,y) = f(x) + \tilde{g}(z) + y^T(z-Ax)+ (\rho /2) \Vert z- Ax \Vert_2^2 $. The $x$-update step involves the minimization of $L_\rho$ over $x$ that gives the zero-gradient condition:
\[
  (\rho A^TA + G^TG)x = G^Tb +A^Ty + \rho A^T z.
\]
It is noted that $(\rho A^TA + G^TG)$ is sparse due to sparsity structures of $G^TG, P^TP$ and $D^TD$ shown in \Cref{fig:GtG_sparsity,fig:PtP_sparsity,fig:DtD_sparsity}. Therefore, we can solve the above linear equation more effectively by exploiting the block-diagonal form of $\rho A^TA + G^TG$. The $z$-update step is to minimize $L_\rho$ over $z$ is to find 
\[
\argmin_z  \quad \tilde{g}(z) + (\rho/2) \Vert z - (Ax - y /\rho) \Vert_2^2 
= \argmin_z  \lambda_{1} h(z_1 ; \mathcal{B}_1) + \lambda_{2} h (z_2; \mathcal{B}_2) + (\rho/2) \left \Vert \begin{bmatrix} z_1 \\ z_2 \end{bmatrix} - \begin{bmatrix} L_1 x - y_1/\rho \\ L_2 x - y_2/\rho \end{bmatrix} \right \Vert_2^2.
\]
As the definition of $h$ is associated with a partition being used, we simply denote $h_1(x) := h(x ;\mathcal{B}_1)$ and $h_2(x) := h(x ; \mathcal{B}_2)$. The above $z$-update step turns into two proximal operators using the separable summation property of $\tilde{g}$ \cite[\S 2]{PB14}:
\[
\begin{bmatrix} \prox_{\lambda_1 h_1/\rho} (L_1x - y_1/\rho) \\ \prox_{\lambda_2 h_2 /\rho}(L_2x - y_2/\rho) \end{bmatrix}.
\]
The functions $h_1$ and $h_2$ all take the form of a composite of weighted-$\ell_q$ and $\ell_2$ norms, expressed as $h(x ; \mathcal{B}) = \sum_{l \in \mathcal{B}} w_l \Vert x_l \Vert_2^q$ given in \eqref{eq:weighted_group_norm}. It is well-known that for $q=1$, we can easily modify the result in \cite[\S 6.5]{PB14} to obtain the proximal operator of $h(x;\mathcal{B})$ as the \emph{weighted block-soft thresholding}. That is, for all $ l \in \mathcal{B}$,
\begin{equation}
(\prox_{\alpha h}(u))_l = \left ( 1 - \alpha w_l/ \Vert u_l \Vert_2  \right )_+  \cdot u_l = \begin{cases} (1-\alpha w_l/\Vert u_l \Vert_2 ) u_l , & \Vert u_l \Vert_2 \geq \alpha w_l, \\
0, & \Vert u_l \Vert_2 < \alpha w_l. 
\end{cases}
\label{eq:prox_hq1}
\end{equation}
For the non-convex case of $q=1/2$, thanks to the analytical form of the proximal operator \cite{HLMQY17} that for all $l \in \mathcal{B}$,
\begin{equation}
(\prox_{\alpha h}(u))_l =  \begin{cases} \left ( \frac{16 \Vert u_l \Vert_2^{3/2} \cos^3(R(u_l))}{3\sqrt{3} \alpha w_l + 16 \Vert u_l \Vert_2^{3/2} \cos^3(R(u_l))} \right )u_l & \Vert u_l \Vert_2 > \frac{3}{2} (\alpha w_l)^{2/3}, \\
0, & \Vert u_l \Vert_2 \leq \frac{3}{2}(\alpha w_l)^{2/3},
\end{cases}
\label{eq:prox_hq12}
\end{equation}
where $R(u_l) = \pi/3 - (1/3)\arccos ( \frac{\alpha w_l}{4}  ( 3/\Vert u_l \Vert_2)^{3/2} )$. The proximal operators \eqref{eq:prox_hq1} and \eqref{eq:prox_hq12} can be computed in parallel for all blocks in the partitions $\mathcal{B}_1$ and $\mathcal{B}_2$. 

The ADMM algorithm for solving \eqref{eq:vec_formulation2} is named \textbf{SparseGrangerNet} and now presented in \Cref{alg:SpectralSparseGrangerNet}. After the ADMM step of updating $x,z$ and $y$ (dual variable), the primal and dual residuals $(r,s)$ are computed. We follow the stopping criterion on these two residuals given in \cite[\S 3.3.1]{BPC11} where the absolute tolerance ($\epsilon_{\mathrm{abs}}$), relative tolerance ($\epsilon_{\mathrm{rel}}$) are set to $10^{-7}$ and $10^{-5}$, respectively. We also implement two choices of $\rho$-update rules for every $T$ iteration because it has been known that the ADMM parameter ($\rho$) greatly affects the algorithm convergence; the two sub-routines are described in \Cref{alg:SADMM,alg:HADMM}. 

For convex formulations, the adaptive rule presented in \Cref{alg:SADMM} follows the spectral penalty selection proposed by \cite{XFG17}. The rule was inspired by the Barzilai-Borwein (BB) gradient method that approximated the Hessian matrix of the objective function in smooth unconstrained problems. The adaptive rule was brought into ADMM in \cite{XFG17} with a safeguard step for measuring a goodness of fit for linear approximations of subgradients of dual ADMM objective that was split into two terms according to the conjugate of $f$ and $g$. The linear approximation of each term was parametrized by two choices of spectral step sizes: steepest descent and minimum gradient, and some hybrid rule further applied to determine the step size. When the linear approximations were sufficiently credible (as measured by correlations), the penalty was updated as the geometric mean or one of the step sizes; otherwise, the previous $\rho$ was kept for the next iteration.

When the standard ADMM with a fixed $\rho$ is used to solve non-convex problems, there is no guarantee for a convergence. It is observed that if $\rho$ is too large, the primal residual iterations have a fast convergence but not for the dual residual; if $\rho$ is too small, the iterations could diverge. In the non-convex case, we then used a heuristic update step described in \Cref{alg:HADMM}; we start $\rho$ with a small value and increase it by a factor of 2 every $T$ iteration. If the primal residual converges, we terminate the update scheme to avoid a slow convergence from $\rho$ being too large. Increasing $\rho$ by a factor greater than one was also proposed for convex problems in \cite{XLLY17} as LA-ADMM, with an improved iteration complexity from a fixed-penalty scheme, where the choice of initial $\rho$ depended on properties of the objective function. Unlike \Cref{alg:HADMM}, the scheme of \cite{XLLY17} has no termination rule; $\rho$ can increase to a large value. 

\begin{algorithm}
\DontPrintSemicolon
\caption{SparseGrangerNet}
parameters: $A = \begin{bmatrix} L_1 \\ L_2 \end{bmatrix},\epsilon_{\mathrm{pri}} ,\epsilon_{\mathrm{dual}},T$ \;
initialization: $ x,y = (y_{1},y_{2}),z=(z_{1},z_{2}), (x,y,z)_{\rm{cached}}, \rho > 0, k=1 $\;
 \While(\tcp*[r]{stopping criterion}){$\Vert r \Vert_{2} \geq \epsilon_{\mathrm{pri}}$ \rm{and} $\Vert s \Vert_{2} \geq \epsilon_{\mathrm{dual}}$}{
$x^{+} =  (\rho A^TA +G^{T}G)^{-1}\left (G^{T}b+ A^T (y+\rho z) \right )$  \tcp*[r]{Exploit block-diagonal $\rho A^TA + G^TG$}  
$z_{1}^{+} = \prox_{\lambda_{1}h_{1}/ \rho}(L_{1}x^{+}-y_{1}/ \rho)$ \tcp*[r]{thresholding with partition $\mathcal{B}_1$}  
$z_{2}^{+} = \prox_{\lambda_{2}h_{2}/ \rho}(L_{2}x^{+}-y_{2}/ \rho)$ \tcp*[r]{thresholding with partition $\mathcal{B}_2$}  
$ y^+ = y + \rho (z^+ - A x^+)$ \;
$ r = z^+- A x^+$  \tcp*[r]{primal residual}  
$ s = \rho A^T (z^+-z)$  \tcp*[r]{dual residual}  
\eIf(\tcp*[r]{Update $\rho$ every $T$ iterations}){$\mod(k,T)=0$}{
$\rho^{+} = \mathrm{UpdatePenalty}(\cdot)$
}
{$\rho^{+} = \rho$}
$k \leftarrow k+1$\;
}
\label{alg:SpectralSparseGrangerNet}
\end{algorithm}

\paragraph{Convergence of Lagrangian-based algorithms for DGN and FGN.} As mentioned earlier, nmAPG or the proximal gradient method are applicable to \eqref{eq:vec_formulation1} with a convergence but it is too computationally demanding because our $g$ in \eqref{eq:penalty_cgn}-\eqref{eq:penalty_fgn} is not proximal-friendly. After re-arranging the formulations as in \eqref{eq:ADMMformat}, the structures of DGN and FGN fall into a class of non-convex composite optimization where the two non-smooth terms in the objective are composite functions with linear transformations; each of which has its nice form of proximal operator. Thus, we resort to find algorithms that split the minimization into easier steps like ADMM.  Literature of ADMM convergence for non-convex problems often depends on the property of matrix $A$ and $B$ of the ADMM format \eqref{eq:ADMMformat}. The analysis from \cite{LP15,WCX18} provided convergence proofs using the full row rank assumption on $A$, which in our case, as given in \eqref{eq:admm_AB}, does not hold for DGN and FGN because our $A$ always has a nonzero nullspace, which can be shown as follows. From \eqref{eq:admm_AB} and that $D=\tilde{D}P$, we have $ \nullspace{(P)} \subseteq \nullspace{(A)}$. The projection matrix $P$ always has a nonzero nullspace; thus, $A$ also has a nonzero nullspace and hence is never full rank. Despite the full row rank assumption, a weaker assumption given in \cite{WYZ19} was that $\Range(B) \subset \Range(A)$; however, this assumption still does not hold in our problem because $B$ is full range but $A$ is not. When we explore into broader types of Lagrangian-based algorithms, one of which is ADMM, a unified treatment of convergence analysis was reviewed in \cite{ST19}. A recent \emph{adaptive Lagrangian-based multiplier (ALBUM)} method \cite{ST19} for non-convex composite problems relies on the so called \emph{a uniform regularity} condition of the composite mapping, which essentially says, in our case, that $A$ must be surjective, similar to full rank assumption of $A$ in \cite{LP15}. To the best of our knowledge, the undesirable property of our $A$ has become the main obstacle to analyze a convergence of ADMM when applied to DGN and FGN. We leave this as an open problem, while our implementation (with fine-tuned parameters) of adaptive ADMM (\Cref{alg:SpectralSparseGrangerNet,alg:HADMM}) to DGN and FGN did not return divergent instances in our experiments.

\begin{algorithm}
\DontPrintSemicolon
\caption{UpdatePenalty($\cdot$) for \textbf{convex} formulations: Spectral adaptive $\rho$ \cite{XFG17}}
parameters: $A = \begin{bmatrix} L_1 \\ L_2 \end{bmatrix}, \epsilon_{c}$ \;
 input: $\rho>0$, $x^{+},(x,y,z),(x,y,z)_{\rm{cached}}$\;
$ \hat{y} = y+\rho(z- A x^{+}) , \Delta \hat{y} = \hat{y}-\hat{y}_{\rm{cached}} $\;
$ \Delta F = A(x-x_{\rm{cached}})$  \tcp*[r]{$\Delta$ subdifferential of dual obj. from $f$}
$a_{1} = \frac{\Delta F^{T}\Delta \hat{y}}{\Vert \Delta F \Vert_{2}^{2}} $, $a_{2} = \frac{\Vert \Delta \hat{y} \Vert_{2}^{2}}{\Delta F^{T}\Delta \hat{y}} $ \tcp*[r]{a1: minimum gradient, a2:steepest descent} 
\eIf(\tcp*[r]{choose spectral step size for $\Delta F$} ){$2a_{1}>a_{2}$}{
$a = a_{1}$
}{
$a = a_{2}-0.5a_{1}$\;
}
$\Delta y = y-y_{\rm{cached}}, \Delta G = -(z-z_{\rm{cached}})$ \tcp*[r]{$\Delta$ subdifferential of dual obj. from $g$}
$b_{1} = \frac{\Delta G^{T}\Delta y}{\Vert \Delta G \Vert_{2}^{2}}$, $b_{2} = \frac{\Vert \Delta y \Vert_{2}^{2}}{\Delta G^{T}\Delta y}$ \tcp*[r]{b1: minimum gradient, b2:steepest descent} 
\eIf(\tcp*[r]{choose spectral step size for $\Delta G$} ){$2b_{1}>b_{2}$}{
$b = b_{1}$
}{
$b = b_{2}-0.5b_{1}$\;
}
$c_{1}=\frac{\Delta F^{T}\Delta \hat{y}}{\Vert \Delta F \Vert_{2} \Vert \Delta \hat{y} \Vert_{2}}$,  $c_{2}=\frac{\Delta G^{T}\Delta y}{\Vert \Delta G \Vert_{2} \Vert \Delta y \Vert_{2}}$ \tcp*[r]{correlation terms: linear approximations of $\Delta F, \Delta G$} 
\tcp*[l]{Safeguard update rule for $\rho$} 
  \uIf(\tcp*[r]{use geometric mean when high correlations}){$c_{1}>\epsilon_{c}$ and $c_{2}>\epsilon_{c}$}{
    $\rho^{+} = \sqrt{ab}$ \;
  }
  \uElseIf{$c_{1}>\epsilon_{c}$ and $c_{2}\leq \epsilon_{c}$}{
    $\rho^{+} = a$ \;
  }
  \uElseIf{$c_{1}\leq \epsilon_{c}$ and $c_{2}>\epsilon_{c}$}{
    $\rho^{+} = b$ \;
  }
  \Else{
     $\rho^{+} = \rho$ \;
  }
\label{alg:SADMM}
\end{algorithm}

\begin{algorithm}
\DontPrintSemicolon
\caption{UpdatePenalty($\cdot$) for \textbf{non-convex} formulations} 
input: $\rho>0, \; r ,\; \epsilon_{\mathrm{pri}}$\;
\eIf{$\Vert r \Vert_{2} \geq \epsilon_{\mathrm{pri}}$}{
$\rho^{+} = 2\rho$\\
}{
$\rho^{+} = \rho$\\
} 
\label{alg:HADMM}
\end{algorithm}

\newpage
\section{fMRI data}
\label{sec:fmridata}
The fMRI time-series data were obtained from the ADHD-200 data sets by the ADHD 200 consortium and are available at: \url{https://www.nitrc.org/plugins/mwiki/index.php/neurobureau:AthenaPipeline}. We pre-processed the data using 14 steps according to the Athena functional data processing pipeline~\cite{BCC17} but without the bandpass filtering step ($0.009-0.08$ Hz) since \cite{SHC12} reported that ADHD and TDC (control) groups were highly discriminative when using the cross-spectral density at the frequency around $0.2$ Hz as a feature. The data were collected from NYU site and screened under the criterions: i) the subjects were male adolescents of 7-17 years old ($11.71 \pm 3.11$), ii) the ADHD and TDC groups were age-matched, iii) the subjects had no secondary diagnosis, iv) the subjects are right-handed with a score larger than $0.1$, where the score ranged from -1 to 1 (from left-handed to right-handed), iv) the subjects had verbal IQ in the range of 98-112, and v) ADHD subjects were combined subtype, \ie, ADHD type with the presence of both inattention and hyperactivity/impulsivity. Under these selections, we obtained $18$ subjects for each of ADHD and TDC groups. The regions of interest (ROI) were defined from the AAL atlas~\cite{TLP02} as shown in \Cref{table:AAL_atlas}. 

\begin{table}[ht]
\caption{AAL atlas.}
\label{table:AAL_atlas}
\centering
\begin{tabular}{llllll} \hline
\# & ROI                                       & \# & ROI (continue)                        & \#  & ROI (continue)                        \\ \hline
1  & Precentral gyrus\_L                        & 40 & Parahippocampal gyrus\_R    & 79  & Heschl gyrus\_L             \\
2  & Precentral gyrus\_R                        & 41 & Amygdala\_L                 & 80  & Heschl gyrus\_R             \\
3  & Superior frontal gyrus (dorsolateral)\_L   & 42 & Amygdala\_R                 & 81  & Superior temporal gyrus\_L  \\
4  & Superior frontal gyrus (dorsolateral)\_R   & 43 & Calcarine cortex\_L         & 82  & Superior temporal gyrus\_R  \\
5  & Superior frontal gyrus (orbital)\_L        & 44 & Calcarine cortex\_R         & 83  & Temporal pole (superior)\_L \\
6  & Superior frontal gyrus (orbital)\_R        & 45 & Cuneus\_L                   & 84  & Temporal pole (superior)\_R \\
7  & Middle frontal gyrus\_L                    & 46 & Cuneus\_R                   & 85  & Middle temporal gyrus\_L    \\
8  & Middle frontal gyrus\_R                    & 47 & Lingual gyrus\_L            & 86  & Middle temporal gyrus\_R    \\
9  & Middle frontal gyrus (orbital)\_L          & 48 & Lingual gyrus\_R            & 87  & Temporal pole (middle)\_L   \\
10 & Middle frontal gyrus (orbital)\_R          & 49 & Superior occipital gyrus\_L & 88  & Temporal pole (middle)\_R   \\
11 & Inferior frontal gyrus (opercular)\_L      & 50 & Superior occipital gyrus\_R & 89  & Inferior temporal gyrus\_L  \\
12 & Inferior frontal gyrus (opercular)\_R      & 51 & Middle occipital gyrus\_L   & 90  & Inferior temporal gyrus\_R  \\
13 & Inferior frontal gyrus (triangular)\_L     & 52 & Middle occipital gyrus\_R   & 91  & Cerebellum\_Crus1\_L         \\
14 & Inferior frontal gyrus (triangular)\_R     & 53 & Inferior occipital gyrus\_L & 92  & Cerebellum\_Crus1\_R         \\
15 & Inferior frontal gyrus (orbital)\_L        & 54 & Inferior occipital gyrus\_R & 93  & Cerebellum\_Crus2\_L         \\
16 & Inferior frontal gyrus (orbital)\_R        & 55 & Fusiform gyrus\_L           & 94  & Cerebellum\_Crus2\_R         \\
17 & Rolandic operculum\_L                      & 56 & Fusiform gyrus\_R           & 95  & Cerebellum\_3\_L             \\
18 & Rolandic operculum\_R                      & 57 & Postcentral gyrus\_L        & 96  & Cerebellum\_3\_R             \\
19 & Supplementary motor area\_L                & 58 & Postcentral gyrus\_R        & 97  & Cerebellum\_4\_5\_L          \\
20 & Supplementary motor area\_R                & 59 & Superior parietal gyrus\_L  & 98  & Cerebellum\_4\_5\_R          \\
21 & Olfactory cortex\_L                        & 60 & Superior parietal gyrus\_R  & 99  & Cerebellum\_6\_L             \\
22 & Olfactory cortex\_R                        & 61 & Inferior parietal gyrus\_L  & 100 & Cerebellum\_6\_R             \\
23 & Superior frontal gyrus (medial)\_L         & 62 & Inferior parietal gyrus\_R  & 101 & Cerebellum\_7b\_L            \\
24 & Superior frontal gyrus (medial)\_R         & 63 & Supramarginal gyrus\_L      & 102 & Cerebellum\_7b\_R            \\
25 & Superior frontal gyrus (medial orbital)\_L & 64 & Supramarginal gyrus\_R      & 103 & Cerebellum\_8\_L             \\
26 & Superior frontal gyrus (medial orbital)\_R & 65 & Angular gyrus\_L            & 104 & Cerebellum\_8\_R             \\
27 & Rectus gyrus\_L                            & 66 & Angular gyrus\_R            & 105 & Cerebellum\_9\_L             \\
28 & Rectus gyrus\_R                            & 67 & Precuneus\_L                & 106 & Cerebellum\_9\_R             \\
29 & Insula\_L                                  & 68 & Precuneus\_R                & 107 & Cerebellum\_10\_L            \\
30 & Insula\_R                                  & 69 & Paracentral lobule\_L       & 108 & Cerebellum\_10\_R            \\
31 & Anterior cingulate gyrus\_L                & 70 & Paracentral lobule\_R       & 109 & Vermis\_1\_2                \\
32 & Anterior cingulate gyrus\_R                & 71 & Caudate\_L                  & 110 & Vermis\_3                   \\
33 & Median cingulate gyrus\_L                  & 72 & Caudate\_R                  & 111 & Vermis\_4\_5                \\
34 & Median cingulate gyrus\_R                  & 73 & Putamen\_L                  & 112 & Vermis\_6                   \\
35 & Posterior cingulate gyrus\_L               & 74 & Putamen\_R                  & 113 & Vermis\_7                   \\
36 & Posterior cingulate gyrus\_R               & 75 & Pallidum\_L                 & 114 & Vermis\_8                   \\
37 & Hippocampus\_L                             & 76 & Pallidum\_R                 & 115 & Vermis\_9                   \\
38 & Hippocampus\_R                             & 77 & Thalamus\_L                 & 116 & Vermis\_10                  \\
39 & Parahippocampal gyrus\_L                   & 78 & Thalamus\_R                 &     &             \\ \hline                
\end{tabular}
\end{table}


\newcommand{\etalchar}[1]{$^{#1}$}

\end{document}